\documentclass{tlp}
\usepackage{aopmath}

\usepackage{nomencl}
\let\abbrev\nomenclature
\renewcommand{\nomname}{List of Symbols and Abbreviations}
\makenomenclature
\newcommand{\Listofabbrev}{
\printnomenclature
\newpage
}

\long\def\comment#1{}

\usepackage{latexsym}
\usepackage{amssymb}

\newtheorem{definition}{Definition} % [section]
\newtheorem{example}{Example} % [section]

\newenvironment{thm*}[2]%
{\begin{trivlist} \item[] {\it #1~\protect{\ref{#2}}} (see
page~\pageref{#2})}{\end{trivlist}}

\newcommand{\nop}[1]{}

\newlength{\dummy}

\def\bldl{\smallskip\[\bf\begin{array}{ll}}
\def\cldl{\vspace{-0.4cm}\[\bf\begin{array}{ll}}
\def\eldl{\end{array}\]\rm}
\def\okay{{\it okay}}

\def\JFP{{\rm JFP}}

\def\fail{{\it fail}}
\def\naf{${\em not~}$}

\def\And{{\wedge}}
\def\gp{{\cal P}}

\def\DLP{\mbox{DLP}^{\mathcal{A}}}
\def\LPA{\mbox{LP}^{\mathcal{A}}}

\def\And{{\wedge}}

\submitted {16 January 2007 }

\revised {13 November 2007, 18 April 2008}

\accepted {21 May 2008}

\begin{document}
\bibliographystyle{acmtrans}

\title[A Knowledge-Representation Language with Social Features]
{Logic Programming with Social Features\footnote{An abridged version of this paper appears in
\cite{BuccaLPNMR05}.}}
\author[F. Buccafurri and G. Caminiti]
{FRANCESCO BUCCAFURRI and GIANLUCA CAMINITI\\
DIMET - Universit\`a ``Mediterranea'' degli Studi di Reggio Calabria\\
via Graziella, loc. Feo di Vito, 89122 Reggio Calabria, Italia\\
\email{bucca@unirc.it, gianluca.caminiti@unirc.it}}

\maketitle

\label{firstpage}

\begin{abstract}

In everyday life it happens that a person has to reason about what other people think
and how they behave, in order to achieve his goals. In other words, an individual may
be required to adapt his behaviour by reasoning about the others' mental state. In
this paper we focus on a knowledge representation language derived from logic
programming which both supports the representation of mental states of individual
communities and provides each with the capability of reasoning about others' mental
states and acting accordingly. The proposed semantics is shown to be translatable
into stable model semantics of logic programs with aggregates. To appear in Theory
and Practice of Logic Programming (TPLP).

\end{abstract}


\section{Introduction}\label{sect:introduction}

In everyday life it happens that a person has to reason about what other people think
and how they behave, in order to achieve his goals. In other words, an individual may
be required to adapt his behaviour by reasoning about the others' mental state. This
typically happens in the context of cooperation and negotiation: for instance, an
individual can propose his own goals if he knows that they would be acceptable to the
others. Otherwise he can decide not to make them public. As a consequence, one can
increase the success chances of his actions, by having information about the other
individuals' knowledge.

In this paper we focus on a knowledge representation language derived from logic
programming which both supports the representation of mental states of individual
communities and provides each with the capability of reasoning about others' mental
states and acting accordingly. The proposed semantics is shown to be translatable
into stable model semantics of logic programs with aggregates.

We give the flavor of the proposal by two introductory examples,
wherein we describe the features of our approach in an informal,
yet deep fashion. Even though in these examples, as well as
elsewhere in the paper, we use the term {\em agent} to denote the
individual reasoning, we remark that our focus is basically
concerning to the knowledge-representation aspects, with no
intention to investigate how this {\em reasoning layer} could be
exploited in the intelligent-agent contexts.
However, in Section~\ref{sect:relwork}, we relate our work with some conceptual
aspects belonging to this research field.

Consider now the first example.

\begin{example}\label{exa:wedding}
There are four agents which have been invited to the same wedding party. These
are the desires of the agents:
\begin{description}
\item[{\bf Agent$_1$}] will go to the party only if at least the half of the total
number of agents (not including himself) goes there.

\item[{\bf Agent$_2$}] possibly does not go to the party, but he tolerates such an
option. In case he goes, then he possibly drives the car.

\item[{\bf Agent$_3$}] would like to join the party together with {\bf Agent$_2$},
but he does not trust on {\bf Agent$_2$}'s driving skill. As a consequence, he
decides to go to the party only if {\bf Agent$_2$} both goes there and does not want
to drive the car.

\item[{\bf Agent$_4$}] does not go to the party.
\end{description}

Now, assume that some agents are less autonomous than the others, i.e. they may
decide either to join the party or not to go at all, possibly depending on the other
agents' choice. Moreover some agents may not require, yet tolerate some options.
\end{example}

The standard approach to representing communities by means of logic-based
agents~\cite{DBLP:conf/atal/SatohY02,dalijelia02,laimadalt05,socs04,subrahmanian00heterogeneous}
is founded on suitable extensions of logic programming with negation as failure ({\em
not}) where each agent is represented by a single program whose intended models
(under a suitable semantics) are the agent's desires/requests. Although we take this
as a starting point, it is still not suitable to model the above example because of
two following issues:

\begin{enumerate}

\item
There is no natural representation for tolerated options, i.e. options which are
not requested, but possibly accepted (see {\bf Agent$_2$}).

\item A machinery is missing which enables one agent to reason about the behaviour of
other agents (see {\bf Agent$_1$} and {\bf Agent$_3$}).

\end{enumerate}

In order to solve the first issue (item 1.) we use an extension of standard
logic programming exploiting the special predicate $okay()$, previously
introduced in \cite{BucGot02}. Therein a model-theoretic semantics aimed to
represent a common agreement in a community of agents was given. However,
representing the requests/acceptances of single agents in a community is not
enough. Concerning item 2 above, a social language should also provide a
machinery to model possible interference among agents' reasoning (in fact it is
just such an interference that distinguishes the social reasoning from the
individual one).
To this aim, we introduce a new construct providing one agent with the ability
to reason about other agents' mental state and then to act accordingly.

Program rules have the form:
\begin{center}
$
\begin{array}{c}
  \mathit{head} \leftarrow [\mbox{\em selection\_condition}] \{\mathit{body}\},
\end{array}
$
\end{center}
where {\em selection\_condition} predicates about some {\em social condition} concerning
either the cardinality of communities or particular individuals satisfying {\em
body}.

For instance, consider the following rule, belonging to a program representing a
given agent $A$:
\begin{center}
$
\begin{array}{c}
a \leftarrow [ \: \mathit{l}, \mathit{h} \: ] \: \{b, \ \naf \: c\}
\end{array}
$
\end{center}

This rule means that agent $A$ will require $a$ in case a number $\nu$ of agents
(other than $A$) exists such that they require or tolerate $b$, neither require nor
tolerate $c$ and it holds that $0 \leq l \leq \nu \leq h \leq n-1$. By default, $l =
0$ and $h = n-1$. The number $n$ is a parameter -- known by each agent --
representing the total number of agents (including the agent $A$). This enriched
language is referred to as {\em SOcial Logic Programming} (SOLP).
\abbrev{SOLP}{SOcial Logic Programming} The wedding party scenario of
Example~\ref{exa:wedding} can be represented by the four SOLP programs shown in Table
\ref{tab:wedding example}, where the program $\gp_4$ is empty since
the corresponding agent has not any request or desire to express.
\begin{table}
\caption{The wedding party (Example~\ref{exa:wedding})} \label{tab:wedding
example}

\programmath $
{\small
\begin{array}{lrl}
\hline\hline \gp_1~(\mbox{\bf Agent$_1$}): &
 go\_wedding \leftarrow & [\frac{n}{2} - 1$,$ \ ]\{go\_wedding\}\\
\hline \gp_2~(\mbox{\bf Agent$_2$}): &
\okay(go\_wedding) \leftarrow & \\
& \okay(drive) \leftarrow & go\_wedding \\
\hline \gp_3~(\mbox{\bf Agent$_3$}): &
go\_wedding \leftarrow & [\mbox{\bf Agent$_2$}]\{go\_wedding$,$ \ \naf drive\}\\
\hline \gp_4~(\mbox{\bf Agent$_4$}): &
\multicolumn{2}{c}{\mbox{empty program}}\\
\hline\hline
\end{array}
}
$ \unprogrammath
\end{table}

The intended models represent the mental states of each agent inside the
community. Concerning the party, such models are the following:

$M_1=\emptyset$, $M_2=\{ go\_wedding_{\gp_1}, go\_wedding_{\gp_2}, drive_{\gp_2}\}$,
and $M_3=\{go\_wedding_{\gp_1},$ $ go\_wedding_{\gp_2}, go\_wedding_{\gp_3}\}$,
where the subscript $_{\gp_i}$ $(1 \leq i \leq n)$ references, for each atom in
a model, the program (resp. agent) that atom is entailed by. The models
respectively mean that either $(M_1)$ no agent will go to the party, $(M_2)$
only {\bf Agent$_1$} and {\bf Agent$_2$} will go and also {\bf Agent$_2$} will
drive the car, or $(M_3)$ all agents but {\bf Agent$_4$} will go to the party.

Let us show why the above models represent the intended meaning of the program: $M_1$
is empty in case {\bf Agent$_2$} does not go to the wedding party (i.e. $go\_wedding$
is not derived by $\gp_2$). Indeed, in such a case, {\bf Agent$_3$} will not go too,
since his requirements w.r.t. {\bf Agent$_2$} are not satisfied. Moreover, since {\bf
Agent$_4$} expresses neither requirements nor tolerated options, he does not go to
the party (observe that such a behaviour is also represented by the models $M_2$ and
$M_3$). Finally, {\bf Agent$_1$} requires that at least one\footnote{Since
$\frac{n}{2}-1=1$, where $n$ is the total number of agents, i.e. $n$=4.} agent (other
than himself) goes to the party. As a consequence of the other agents' behaviour,
{\bf Agent$_1$} will not go. Thus, no agent will go to the party and $M_1$ is empty.
The intended meaning of $M_2$ is that both {\bf Agent$_1$} and {\bf Agent$_2$} will
go to the party and {\bf Agent$_2$} will also drive the car. In such a case {\bf
Agent$_3$} will not go since he requires that {\bf Agent$_2$} does not drive the car.
The model $M_3$ represents the case in which {\bf Agent$_2$} goes to the party, but
does not drive the car. Now, since all requirements of {\bf Agent$_3$} are satisfied,
then he also will go to the party. Certainly, {\bf Agent$_1$} will join the other
agents, because, in order to go to the party, he requires that at least one agent
goes there.

The intended models are referred to as {\em social models}, since they express the
results of the interactions among agents. As it will be
analyzed in Section~\ref{sect:complexity}, the multiplicity of intended models is
induced both by negation occurring in rule bodies and also directly by the social
features, thus making the approach non-trivial.

Let us informally introduce the most important properties of the semantics of the
language:

\begin{itemize}
\item Social conditions model reasoning conditioned by the behaviour of other agents
in the community. In particular, it is possible to represent collective mental
states, preserving the possibility of identifying the behaviour of each agent.

\item It is possible to nest social conditions, in order to apply recursively the
social-conditioned  reasoning to agents' subsets of the community.

\item Each social model represents the mental state (i.e. desires, requirements,
etc.) of every agent in case the social conditions imposed by the agents are enabled.
\end{itemize}

Observe that, in order to meet such goals, merging all the input
SOLP programs it is not enough, since this way we lose all
information about the relationship between an atom and the program
(resp. agent) which such an atom comes from. Therefore, we have to
find a non-trivial solution.

Our approach starts from~\cite{BucGot02}, where the {\em Joint Fixpoint Semantics}
(JFP)\abbrev{JFP}{Joint Fixpoint Semantics}, that is a semantics providing a way to
reach a compromise (in terms of a common agreement) among agents modelled by logic
programs, is proposed. Therein, each model contains atoms representing items being
common to all the agents. The approach proposed here in order to reach a social-based
conclusion is more general: the agents' behaviour is defined by taking into account
social conditions specified by the agents themselves.

Informally, a social condition is an expression $[\mbox{\em selection\_condition}]
\{\mathit{body}\}$, where {\em selection\_condition} can be of two forms: either (i)
cardinality-based, or (ii) identity-based. In the former case the agent requires that
a number of other agents (bounded by {\em selection condition} itself) satisfy {\em
body}. In the latter case, {\em selection condition} identifies which agent is
required to satisfy {\em body}. Given a program rule including a social condition
such as $\mathit{head} \leftarrow [\mbox{\em selection\_condition}]
\{\mathit{body}\}$, the intuitive meaning is that $\mathit{head}$ is derived if the
social condition is satisfied.

An example of cardinality-based condition (case (i) above) is shown in
Table~\ref{tab:wedding example} by the program $\gp_1$: an intended model $M$ will
include the atom $go\_wedding_{\gp_1}$ if a set of programs $S' \subseteq
\{\gp_2,\gp_3,\gp_4\}$ exists such that for each $\gp \in S'$, it results that
$go\_wedding_{\gp}$ belongs to $M$ and also it holds that the number of programs in
$S'$ satisfies the social condition imposed by the program $\gp_1$, that is $|S'|
\geq \programmath \frac{n}{2}-1 \unprogrammath $.

An example of case (ii) (identity-based condition) is represented by the program
$\gp_3$, which requests the atom $go\_wedding_{\gp_3}$ to be part of an intended
model $M$ if $go\_wedding_{\gp_2}$ belongs to $M$, but the atom $drive_{\gp_2}$ does
not. Importantly, social conditions can be nested each other, as shown by the next
example.

\begin{example}\label{exa:download}
Consider a Peer-to-Peer file-sharing system where a user can share his collection of
files with other users on the Internet. In order to get better performance, a file is
split in several parts being downloaded separately (possibly each part from a
different user)\footnote{Among others, KaZaA, EMule and BitTorrent are the most
popular Internet P2P file-sharing systems exploiting such a feature.}. The following
SOLP program $\gp$ describes the behaviour of an agent (acting on behalf of a given
user) that wants to download every file $X$ being shared by at least a number
$\mathit{min}$ of users such that at least one of them owns a complete version of $X$
(rule $r_1$). Moreover, the agent tolerates to share any file $X$ of his, which is
shared also by at least the 33\% of the total number of users in the network and such
that among those users, a number of them (bounded between 20\% and 70\% of the total)
exists having a high bandwidth. In this case the agent tolerates to share, since he
is sure that the network traffic will not be unbalanced (rule $r_2$). Observe that
the use of nested social conditions in $\gp$ is emphasized by means of program
indentation.

\begin{center}
$
{\small
\begin{array}{rrl}
r_1: & download(X) \leftarrow & [min, ]\{share(X), \\
    &               & \qquad [1, ]\{\naf incomplete(X)\\
    &               &  \}, file(X)\\
r_2: & okay(share(X)) \leftarrow & [0\mbox{.}33*n, ]\{share(X), \\
    &               & \qquad [0\mbox{.}2*n, 0\mbox{.}7*n]\{high\_bw\} \\
    &               & \}, file(X)\\
\end{array}
}
$
\end{center}
\end{example}

Now, one could argue that a different choice concerning the selection condition could
be done. As a first observation we note that the chosen selection conditions play
frequently an important role in common-sense reasoning. Indeed, it often happens that
the beliefs and the choices  of an individual depend on {\em how many} people think
or act in a certain way. For instance, a person who needs a new mobile phone is
interested in collecting -- from his colleagues or the Internet -- a number of
opinions on a given model, in order to decide whether he should buy it. It occurs
also that one is interested in the behaviour of a {\em given} person, in order to act
or to infer something. For example, two people are doing shopping together and do not
want to buy the same clothes in order not to be dressed the same way. So, one of them
decides not to buy a given item, in case it has been chosen by his partner. These
short examples show that two important parameters acting in the social influence are
either (i) the number or (ii) the identity of the people involved. For such a reason,
we propose a simple, clear-cut, yet general mechanism to represent the selection of a
social condition.

As a second observation we remark that this work represents a first step towards a
thorough study on how to include in a classical logic-programming setting the
paradigm of social interference, in order to directly represent community-based
reasoning. To this aim, we focus on some suitable selection conditions, but we are
aware that other possible choices might be considered. From this perspective, our
work tries to give some non-trivial contributions towards what kind of features a
knowledge representation language should include, in order to be oriented to complex
scenarios. Anyway, as it will be shown by examples throughout the paper, the chosen
social conditions combined with the power of nesting allow us to represent more
articulated selections among agents.

Besides the definition of the language, of its semantics and the application to
Knowledge Representation, another contribution of the paper is the translation of
SOLP programs into logic programs with aggregates\footnote{As it will be shown in
Section~\ref{sect:translation}, we choose the syntax of the non-disjunctive fragment
of $\DLP$~\cite{DLPA03}, supported by the DLV system~\cite{Leone02}.}. In particular,
given a set of SOLP programs, a source-to-source transformation exists which provides
as output a single logic program with aggregates whose stable models are in
one-to-one correspondence with the social models of the set of SOLP programs. The
translation to logic programs with aggregates give us the possibility of exploiting
existing engines to compute logic programs.

Moreover, Section~\ref{sect:complexity} shows that our kind of
social reasoning is not trivial, since even in the case of positive programs, the
semantics of SOLP has a computational complexity which is NP-complete.

The paper is organized as follows: in Sections~\ref{sect:syntax} and
\ref{sect:semantics} we define the notion of SOLP programs and their semantics
({\em Social Semantics}), respectively. In Section~\ref{sect:translation} we
illustrate how a set of SOLP programs, each representing a different agent, is
translated into a single logic program with aggregates whose stable models
describe the mental states of the whole agent community and then we show that
such a translation is sound and complete. In Section
\ref{sect:social_models_jfp} we prove that the Social Semantics extends the JFP
Semantics~\cite{BucGot02} and in Section~\ref{sect:complexity} we study the
complexity of several interesting decision problems. Section \ref{sect:kr}
describes how this novel approach may be used for knowledge representation by
means of several examples. Then, in Section~\ref{sect:relwork} we discuss
related proposals and, finally, we draw our conclusions ans sketch the future
directions of the work.

In order to improve the overall readability, these sections are followed
by~\ref{app:proofs} -- where we have placed the proofs of the most complicated
technical results -- and by the list of symbols and abbreviations used
throughout the paper.

\section{Syntax of SOLP Programs}\label{sect:syntax}
In this section we first introduce the notion of {\em social condition} and then we
describe the syntax of SOLP programs.

A {\em term} is either a variable or a constant. Variables are denoted by strings
starting with uppercase letters, while those starting with lower case letters denote
constants. An {\em atom} or {\em positive literal} is an expression  $p(t_1, \cdots,
t_n)$, where $p$ is a {\em predicate} of arity $n$ and $t_1, \cdots, t_n$ are terms.
A {\em negative literal} is the {\em negation as failure (NAF)}\abbrev{NAF}{Negation
As Failure} $\naf a$ of a given atom $a$.

\begin{definition}\label{def:SC}
\abbrev{($n$-)SC}{($n$-)Social Condition} Given an integer $n > 0$, a ($n$-){\em
social condition} $s$, also referred to as ($n$-)SC, is an expression of the form {$
\mathit{cond}(s) \: \mathit{property}(s) $}, such that:
\begin{description}
\item[](1) $\mathit{cond}(s)$
is an expression $[ \alpha ]$ where $\alpha$ is either (i) a pair of integers $l,h$ such that $0
\leq l \leq h \leq n-1$, or (ii) a string;

\item[] (2) $\mathit{property}(s)= \mathit{content}(s) \cup \mathit{skel}(s)$, where
$\mathit{content}(s)$ is a non-empty set of literals and $\mathit{skel}(s)$ is a
(possibly empty) set of SCs.
\end{description}

\end{definition}

$n$-social conditions operate over a collection of $n$ programs representing
the agent community. Each agent is modelled by a program (we will formally
define later in this section which kind of programs are allowed). $n$
represents the total number of agents. In the following, whenever the context
is clear, $n$ is omitted.

Concerning item (1) of the above definition, in case (i), $\mathit{cond}(s)$ is
referred to as {\em cardinal selection condition}, while, in case (ii),
$\mathit{cond}(s)$ is referred to as {\em member selection condition}.

Concerning item (2) of Definition \ref{def:SC}, if $\mathit{skel}(s)=
\emptyset$ then $s$ is said {\em simple}. For a simple SC $s$ such that
$\mathit{content}(s)$ is singleton, the enclosing braces can be omitted.
Finally, given a SC $s$, the formula $\naf s$ is referred to as the NAF of $s$.
The following are {\em simple} SCs extracted from our initial wedding party
example (see Table~\ref{tab:wedding example}):
\begin{example}
\programmath $
 [\frac{n}{2} - 1$,$ \ ]\{go\_wedding\}$.

\unprogrammath
\end{example}
\begin{example}
\programmath

$ [\mbox{\bf Agent$_2$}]\{go\_wedding$,$ \ \naf \ drive\}$.

\unprogrammath
\end{example}

The social conditions occurring in the example regarding a Peer-to-Peer system
(see Example~\ref{exa:download}) are not simple. As a further example, consider
a SC
$s =$ {
$ [l,h]\{a,b,c,[l_1,h_1]\{d,[l_2,h_2]e\},[l_3,h_3]f\}$}. Observe that
$s$ is not simple, since $\mathit{skel}(s)=$ {
 $\{[l_1, h_1] \{d, [l_2, h_2]
e\}, [l_3, h_3] f\}$}, moreover, $\mathit{content}(s)=$ {
 $\{a,b,c\}$}.

Social conditions enable agents to specify requirements over either individual or
groups within the agent community, by using member or cardinal selection conditions,
respectively. Moreover, by nesting social conditions it is possible to declare
requirements over sub-groups of agents, provided that a super-group satisfying a SC
exists. In order to guarantee the correct specifications of nested social conditions,
the notion of {\em well-formed} social condition is introduced next.

Given two n-SCs $s$ and $s'$ such that $\mathit{cond}(s)=[l,h]$ and
$\mathit{cond}(s')=[l',h']$ ($0 \leq l \leq h \leq n-1$, $0 \leq l' \leq h'
\leq n-1$), if $h' \leq h$, then we write $\mathit{cond}(s') \subseteq
\mathit{cond}(s)$.

A SC $s$ is {\em well-formed} if either (i) $s$ is simple, or (ii) $s$ is not simple,
$\mathit{cond}(s)$ is a cardinal selection condition and {$\forall s' \in
\mathit{skel}(s)$} it holds that either (a) $\mathit{cond}(s')$ is a cardinal
selection condition, $s'$ is a well-formed social condition and ${\mathit{cond}(s')
\subseteq \mathit{cond}(s)}$, or (b) $\mathit{cond}(s')$ is a member selection
condition and $s'$ is simple.

According to the intuitive explanation of the above definition,
it results that, besides simple SCs,
only non-simple SCs with cardinal selection condition are candidate to
be well formed. Indeed,
a non-simple SC with member selection condition
requires some property
on a single target agent, but no further sub-group of agents could
be specified by means of SCs possibly nested in it.
Anyway, a further property is required to
SCs with cardinal selection condition in order to be well-formed.
In particular, given a non-simple SC $s$
(with cardinal selection condition), all the SCs nested in
$s$ with cardinal condition must not
exceed the cardinality constraints expressed by $cond(s)$.

\begin{example}\label{exa:well-formed_SC}
The SC $s=$ {
$[1,8]\{a,[3,6]\{b,[{\bf {Agent}_2}]\{c,d\}\}\}$} is well-formed. Note that the
non-simple SCs
$s_1=[{\bf {Agent}_3}]\{a,[3,6]\{b\}\}$ and $s_2=$ {
 $[4,7]\{a,[3,9]b\}$} are not well-formed, because $cond(s_1)$
  is a member selection condition and, concerning $s_2$,
$[3,9]b \in skel(s_2)$ and
  {
 $[3,9] \not\subseteq
[4,7]$}.
\end{example}

From now on, we consider only well-formed SCs.

We introduce now the notion of rule. Our definition generalizes the notion of
classical logic rule.

\begin{definition}
Given an integer $n > 0$, a ($n$-){\em social rule} $r$ is a
formula $a \leftarrow b_1\: \And\cdots\And \: b_m \: \And \: s_1\:
\And\cdots\And \: s_k$ ($m\geq 0$, $k \geq 0$), where $a$ is an
atom, each $b_i$ $(1 \leq i \leq m)$ is a literal and each $s_j$
$(1 \leq j \leq k)$ is either a $n$-SC or the NAF of a $n$-SC.
\end{definition}

Concerning the above definition, the atom $a$ is referred to as the {\em head} of
$r$, while the conjunction $b_1\: \And\cdots\And \: b_m \: \And \: s_1\:
\And\cdots\And \: s_k$ is referred to as the {\em body} of $r$.

In case $a$ is of the form $\mathit{okay}(p)$, where $p$ is an atom, then $r$ it is
referred to as
($n$-){\em tolerance (social) rule}.
In case $k=0$, then a social non-tolerance rule is referred to as {\em classical
rule}.

Social tolerance rules, i.e. rules with head of the form $\mathit{okay}(p)$, encode
tolerance about the occurrence of $p$. The rule $\mathit{okay}(p) \leftarrow
\mathit{body}$ differs from the rule $p \leftarrow \mathit{body}$ since the latter
produces the derivation of $p$ whenever $body$ is satisfied, thus encoding something
that is {\em required} under the condition expressed by $body$. According to the
former rule ($\mathit{okay}(p) \leftarrow \mathit{body}$), the truth of $body$ does
not necessarily imply $p$, yet  its derivation is not in contrast with the intended
meaning of the rule itself. In this sense, under the condition expressed by $body$,
$p$ is just {\em tolerated}.

Given a rule $r$, we denote by $\mathit{head}(r)$ (resp. $\mathit{body}(r)$) the head
(resp. the body) of $r$. Moreover, $r$ is referred to as a {\em fact} in case the
body is empty, while $r$ is referred to as an {\em integrity constraint} if the head
is missing.

\begin{example}
An example of non-tolerance social rule is $a \leftarrow b, c,
[1,9]\{b,c, \naf g, [1,4]\{d\}\},[\gp_2]\{d\}$. An example of tolerance social rule is
$\mathit{okay}(a) \leftarrow \naf b, c, [1,6]\{a, \naf f,g\}, \naf [\gp_2]\{d\}$.
\end{example}

\begin{definition}
A SOLP {\em collection} is a set $\{\gp_1,\cdots, \gp_n\}$ of SOLP programs, where
each SOLP {\em program} is a set of $n$-social rules.
\end{definition}

A SOLP program is {\em positive} if no NAF symbol $not$ occurs in it. For the sake of
presentation we refer, in the following sections, to {\em ground} (i.e.,
variable-free) SOLP programs -- the extension to the general case is straightforward.

\section{Semantics of SOLP programs}\label{sect:semantics}

In this section we introduce the {\em Social Semantics}, i.e. the semantics of a
collection of SOLP programs. We assume that the reader is familiar with the basic concepts
of logic programming \cite{GelLif91,Baral03}.

We start by introducing the notion of interpretation for a single SOLP program. An
{\em interpretation} for a ground (SOLP)\footnote{We insert SOLP into brackets since
the definition is the same as for traditional logic programs.} program $\gp$ is a
subset of $Var(\gp)$, where $Var(\gp)$ is the set of atoms appearing in
$\gp$.\abbrev{$Var(\gp)$}{The set of atoms appearing in $\gp$} A positive literal $a$
(resp.\ a negative literal $\naf a$) is {\em true} w.r.t.\ an interpretation $I$ if
$a\in I$ (resp.\ $a \notin I$); otherwise it is {\em false}. A rule is {\em true}
w.r.t.\ $I$ if its head is true or its body is false w.r.t.\ $I$.

Recall that, for each traditional logic program $Q$, the {\em immediate consequence
operator} $T_Q$\label{def:Tp} is a function from $2^{Var(Q)}$ to $2^{Var(Q)}$ defined
as follows. For each interpretation $I\subseteq Var(Q)$, $T_Q(I)$ consists of the set
of all heads of rules in $Q$ whose bodies are true w.r.t. $I$. An interpretation $I$
is a {\em fixpoint} of a logic program $Q$ if $I$ is a fixpoint of the associated
transformation $T_Q$, i.e., if $T_Q(I)=I$.\abbrev{$T_\gp()$}{The immediate
consequence operator, applied to the program $\gp$}

The set of all fixpoints of $Q$ is denoted by $FP(Q)$\label{def:FP}.
\abbrev{$FP(\gp)$}{The set of all fixpoints of $\gp$}

Before defining the intended models of our semantics, we need some preliminary
definitions.

Let $\gp$ be a SOLP program. We define the {\em autonomous reduction} of $\gp$,
denoted by $A(\gp)$\label{def:A}, the program obtained from $\gp$ by removing all the
SCs from the rules in $\gp$.\abbrev{$A()$}{The autonomous reduction operator}
The intuitive meaning is that in case the program $\gp$ represents the social
behaviour of an agent, then $A(\gp)$ represents the behaviour of the same agent in
case he decides to operate independently of the other agents.

\begin{definition}\label{def:ATp}

\glossary{$AT_\gp()$}{Autonomous immediate consequence operator, applied to the SOLP
program $\gp$} Given a SOLP program $\gp$ and an interpretation $I\subseteq
Var(A(\gp))$, let $TR(A(\gp)$) be the set of tolerance rules in $A(\gp)$ and
$Var^*(A(\gp))$ be the set $Var(A(\gp)) \setminus \{\mathit{okay}(p) \mid
\mathit{okay}(p)\in Var(A(\gp))\} \ \cup \ \{p \mid \mathit{okay}(p)\in
Var(A(\gp))\}$. The {\em autonomous immediate consequence operator} $AT_\gp$ is the
function from $2^{Var^*(A(\gp))}$ to $2^{Var^*(A(\gp))}$, defined as follows:
\abbrev{$TR(\gp)$}{The set of tolerance rules in the program $\gp$}

\begin{tabbing}
$AT_\gp(I) =$ \= $\{ \mathit{head}(r) \mid r \in A(\gp)\setminus TR(A(\gp))
 \ \And \ \mathit{body}(r) \mbox { is true w.r.t. } I \} \ \cup$ \\
 \> $\{ a \mid \mathit{head}(r) = \mathit{okay}(a) \ \And \ r \in TR(A(\gp)) \ \And \
(\mathit{body}(r) \: \And \: a) \mbox { is true w.r.t. } I \}$.\\
\end{tabbing}
\end{definition}

Observe that $AT_\gp$, when applied to an interpretation $I$, extends the classical
immediate consequence operator $T_\gp$, by collecting not only heads of non-tolerance
rules whose  body
is true w.r.t. $I$, but also each atom $a$ occurring as {\em okay}($a$) in the head
of some rule such that both $a$ and the rule body are true w.r.t. $I$.

\begin{definition}\label{def:auto_fp}
An interpretation $I$ for a SOLP program $\gp$ is an {\em autonomous fixpoint} of
$\gp$ if $I$ is a fixpoint of the associated transformation $AT_\gp$, i.e. if
$AT_\gp(I)=I$. The set of all autonomous fixpoints of $\gp$ is denoted by $AFP(\gp)$.

\abbrev{$AFP(\gp)$}{The set of all autonomous fixpoints of $\gp$}
\end{definition}

Observe that by means of the autonomous fixpoints of a given SOLP program $\gp$ we
represent the mental states of the corresponding agent, assuming that every social
condition in $\gp$ is not taken into account.

\begin{example}
Consider the following SOLP program $\gp$:

$
\begin{array}{rcl}

\mathit{okay}(a) &  \leftarrow & b, [1, ]\{c\}\\\
b & \leftarrow & [2,4]\{d\} \\
\end{array}
$

It is easy to see that $AFP(\gp)=\{\{b\},\{a,b\}\}$, i.e. the interpretations
$I_1=\{b\}$ and $I_2=\{a,b\}$ are the autonomous fixpoints of $\gp$, since it holds
that $AT_{\gp}(I_1)=I_1$ and $AT_{\gp}(I_2)=I_2$.

\end{example}

\begin{definition}\label{def:labelled_int}
Given a SOLP collection $C=\{\gp_1, \cdots, \gp_n\}$, let $\gp_i$ ($1 \leq i \leq n$)
be a SOLP program of $C$ and $L$ be a set of atoms. The {\em labelled version} of $L$
w.r.t. $\gp_i$, denoted by $(L)_{\gp_i}$  is the set $\{ a_{\gp_i} \mid a \in L \}$.
Each element of $(L)_{\gp_i}$ is referred to as a {\em labelled atom} w.r.t. $\gp_i$.
\end{definition}

\begin{example}
Given a SOLP program $\gp_1$ of a SOLP collection $C$, if $L=\{a,b,c\}$, then
$(L)_{\gp_1}=\{a_{\gp_1},b_{\gp_1},c_{\gp_1}\}$, where the program identifier
 $\gp_1$ indicates the associated SOLP program.
\end{example}

Now we introduce the concept of {\em social interpretation}, devoted to representing
the mental states of the collectivity described by a given SOLP collection and then
we give the definition of truth for both literals and SCs w.r.t. a given social
interpretation. To this aim, the classical notion of interpretation is extended by
means of program identifiers introducing a link between atoms of the interpretation
and programs of the SOLP collection.

\begin{definition}
Let $C=\{\gp_1,  \cdots, \gp_n\}$ be a SOLP collection. A {\em social interpretation}
for $C$ is a set $\bar{I} = (I^1)_{\gp_1} \cup \cdots \cup (I^n)_{\gp_n}$, where
$I^j$ is an interpretation for $\gp_j$ ($1 \leq j \leq n$) and $(I^j)_{\gp_j}$ is the
labelled version of $I^j$ w.r.t. $\gp_j$ (see Definition~\ref{def:labelled_int}).
\end{definition}

\begin{example}
Given $C=\{\gp_1,  \gp_2, \gp_3\}$, $I^1=\{a,b,c\}$, $I^2=\{a,d,e\}$ and
$I^3=\{b,c,d\}$, where $I^j$ is an interpretation for $\gp_j$ $(1 \leq j \leq 3)$,
then $\bar{I} =\{a_{\gp_1}, b_{\gp_1}, c_{\gp_1}, a_{\gp_2}, d_{\gp_2},$ $e_{\gp_2},
b_{\gp_3}, c_{\gp_3}, d_{\gp_3}\}$ is a social interpretation for $C$.
\end{example}

We define now the notion of truth for literals, SCs and social rules, respectively.

Let $C=\{\gp_1,  \cdots, \gp_n\}$ be a SOLP collection. Given a social interpretation
$\bar{I}$ for $C$ and a positive literal $a \in \bigcup_{\gp \in C} Var(\gp)$, $a$
(resp.\ $\naf a$) is {\em true} for $\gp_j$ ($1 \leq j \leq n$) w.r.t.\ $\bar{I}$ if
$a_{\gp_j} \in \bar{I}$ (resp.\ $a_{\gp_j} \notin \bar{I}$); otherwise it is {\em
false}.

Because of the recursive nature of SCs, before giving the definition of truth for a
SC $s$, we introduce a way to identify $s$ (and also every SC nested in $s$)
occurring in a given rule $r$ of a SOLP program $\gp$. To this aim, we first define
 a function which returns, for a given SC, its nesting depth.

Given a SC $s$, we define the function $depth$\label{def:depth} as follows:

\[ depth(s)=
\left\{
\begin{array}{lr}
depth(s')+1 & \mbox{ if } \exists s' \mid s \in \mathit{skel}(s')\\
0 & \mbox{otherwise.}\\
\end{array}
\right.
\]

\label{def:MSC} Given a SOLP program $\gp$, a social rule $r \in \gp$ and an
integer $n \geq 0$, we define the set ${MSC^{\langle \gp,r,n \rangle}=\{s \mid
s \mbox{ is a SC occurring in } r  \ \And \ depth(s) = n \}}$, i.e. the set
including all the SCs having a given depth $n$ and occurring in a social rule
$r$ of a SOLP program $\gp$. Observe that, in case the parameter $n$ is zero,
then $MSC^{\langle \gp,r,0 \rangle}$ denotes the set of SCs as they appear in
the rule $r$ of $\gp$.

\abbrev{$MSC^{\langle \gp,r,n \rangle}$}{The set of all the SCs having a given depth
$n$ and occurring in a social rule $r$ of a SOLP program $\gp$}

\begin{example}
Let {\small $a \leftarrow  [1,8]\{a,[3,6]\{b,[\gp_2]\{c,d\}\}\}, [2,3]\{e,f\}$} be a
rule $r$ in a SOLP program $\gp_1$. Then:

\begin{center}
{
 $
\begin{array}{l}
MSC^{\langle \gp,r,0 \rangle}= \{ \ [1,8]\{a,[3,6]\{b,[\gp_2]\{c,d\}\}\}, [2,3]\{e,f\} \ \},\\
MSC^{\langle \gp,r,1 \rangle}= \{ \ [3,6]\{b,[\gp_2]\{c,d\}\} \ \},\\
MSC^{\langle \gp,r,2 \rangle}= \{ \ [\gp_2]\{c,d\} \ \},\\
MSC^{\langle \gp,r,3 \rangle}= \emptyset\mbox{.}
\end{array}
$}
\end{center}
\end{example}

Given a SOLP program $\gp$, we define the set $ MSC^\gp=\bigcup_{r \in \gp}
MSC^{\langle \gp,r,0 \rangle}$.

$MSC^\gp$ is the set of all the SCs (with depth 0) occurring in $\gp$.
\abbrev{$MSC^\gp$}{The set of all the SCs (with depth 0) occurring in $\gp$}

Now we provide the definition of truth of a SC w.r.t. a given social interpretation
and, subsequently, the definition of truth of a social rule.

\begin{definition}\label{def:SC_sat}

Let $C = \{\gp_1, \cdots, \gp_n \}$ be a SOLP collection, $C' \subseteq C$ and $\gp_j
\in C'$. Given a social interpretation $\bar{I}$ for $C$ and an $n$-SC $s \in
MSC^{\gp_j}$, we say that $s$ is {\em true} for $\gp_j$ in $C'$ w.r.t. $\bar{I}$ if
it holds that either:

$
\begin{array}{ll}
(1) & \mathit{cond}(s) = [\gp_k] \ \And \\
    & \exists \gp_k \in C' \mid \forall a \in \mathit{content}(s),
    \ a \mbox{ is true for $\gp_k$ w.r.t. } \bar{I}, \mbox{ or}\\
(2) & \mathit{cond}(s) = [l,h] \ \And \\
    & \exists D \subseteq C' \setminus \{\gp_j\}
    \mid l \leq |D| \leq h \ \And \\
    & \forall a \in \mathit{content}(s), \forall \gp \in D,
      \ a \mbox{ is true for $\gp$ w.r.t. } \bar{I} \ \And \\
    & \forall s' \in \mathit{skel}(s) \ \exists D' \subseteq D \mid
    s' \mbox{ is true for $\gp_j$ in } D' \mbox{ w.r.t. }  \bar{I},\\
\end{array}
$

\noindent where $l$, $h$ are integers and $\gp_k$ is a SOLP program.

If $C' = C$, then we simply say that $s$ is {\em true} for $\gp_j$ w.r.t.
$\bar{I}$. An $n$-SC not true for $\gp_j$ (in $C'$) w.r.t. $\bar{I}$ is {\em
false} for $\gp_j$ (in $C'$) w.r.t. $\bar{I}$.

Finally, the NAF of a $n$-SC $s$, $\naf s$, is {\em true} (resp. {\em false}) for $\gp_j$ (in
$C'$) w.r.t. $\bar{I}$ if $s$ is false (resp. true) for $\gp_j$ (in $C'$) w.r.t. $\bar{I}$.

\end{definition} Informally, given a SC
$s$ included in $\gp_j$, $s$ is true for $\gp_j$ w.r.t. a social interpretation
$\bar{I}$ if a single SOLP program $\gp_k$ (resp. a set $D$ of SOLP programs not
including $\gp_j$) exists such that all the elements in $\mathit{content(s)}$ are
true for $\gp_k$ w.r.t. $\bar{I}$ (resp. for every program $\gp \in D$ w.r.t. $\bar{I}$,
and such that every element in $\mathit{skel}(s)$ is true for $\gp_j$ w.r.t. $\bar{I}$).
Observe that
the truth of $\mathit{property}(s)$ is possibly defined recursively, since $s$ may
contain nested SCs.

Once the notion of truth of SCs has been defined, we are able to define the notion of
truth of a social rule w.r.t. a social interpretation.

Let $C$ be a SOLP collection and $\gp \in C$. Given a social interpretation $\bar{I}$
for $C$ and a social rule $r$ in $\gp$, the head of $r$ is {\em true} w.r.t.
$\bar{I}$ if either (i) ($\mathit{head}(r)=a) \ \And \ (a$ is true for $\gp$ w.r.t.
$\bar{I}$), or (ii) ($\mathit{head}(r)=\mathit{okay}(a)) \ \And \ (a$ is true for
$\gp$ w.r.t. $\bar{I}$). Moreover, the body of $r$ is {\em true} w.r.t. $\bar{I}$ if
each element of $\mathit{body}(r)$ is true for $\gp$ w.r.t. $\bar{I}$. Finally, the
social rule $r$ is {\em true} w.r.t. $\bar{I}$ if its head is true w.r.t. $\bar{I}$
or its body is false w.r.t. $\bar{I}$.

Given a SOLP collection $\{\gp_1, \cdots, \gp_n\}$, we define the set of {\em
candidate social interpretations} for $\gp_1, \cdots, \gp_n$ as
\[
\mathcal{U}(\gp_1, \cdots, \gp_n) = \left\{ (F^1)_{\gp_1} \cup \cdots \cup (F^n)_{\gp_n} \mid F^i \in
AFP(\gp_i) \ \And \ 1 \leq i \leq n \right\},
\]

\abbrev{$\mathcal{U}(\gp_1, \cdots, \gp_n)$}{The set of all the possible combinations
of autonomous fixpoints of the SOLP programs $\gp_1, \cdots, \gp_n$} where, recall,
$AFP(\gp_i)$ is the set of autonomous fixpoints of the SOLP program $\gp_i$
(introduced in Definition~\ref{def:auto_fp}) and by $(F^i)_\gp$ ($1\leq i \leq n$) we
denote the labelled version of $F^i$ w.r.t. $\gp$ (see
Definition~\ref{def:labelled_int}).

The set $\mathcal{U}(\gp_1, \cdots, \gp_n)$ represents all the configurations
obtained by combining the autonomous (i.e. without considering the social conditions)
mental states of the agents corresponding to the programs $\gp_1, \cdots, \gp_n$.
Each candidate social interpretation is a candidate intended model. The intended
models are then obtained by enabling the social conditions.

Now, we are ready to give the definition of intended model w.r.t. the Social
Semantics.

\begin{definition}\label{def:ST_C}
\abbrev{$ST_{C}$()}{The social immediate consequence operator, applied to the SOLP
collection $C$} Given a SOLP collection $C=\{\gp_1, \cdots, \gp_n\}$ and a social
interpretation $\bar{I}$ for $C$, let $\bar{V}$ be the set $(Var(\gp_1))_{\gp_1} \
\cup \ \cdots \ \cup \ (Var(\gp_n))_{\gp_n}$ and $TR(\gp_i)$ be the set of tolerance
rules of $\gp_i$ ($1 \leq i \leq n$). The {\em social immediate consequence operator}
$ST_{C}$ is a function from $2^{\bar{V}}$ to $2^{\bar{V}}$ defined as follows:

$
\begin{array}{rl}
ST_{C}(\bar{I})= & \{ a_{\gp} \mid \gp \in C \ \And \ r \in \gp \setminus TR(\gp) \
\And \
\mathit{head}(r) = a \ \And \\
 & \  \mathit{body}(r) \mbox{ is true w.r.t } \bar{I}\} \ \cup \\
 & \{a_{\gp} \mid \gp \in C  \ \And \ r \in TR(\gp) \ \And \ \mathit{head}(r) = \mathit{okay}(a) \ \And \\
 & \  a \mbox{ is true for $\gp$ w.r.t } \bar{I} \ \And \ \mathit{body}(r) \mbox{ is true w.r.t } \bar{I}\}\mbox{.}\\
\end{array}
$

A candidate social interpretation $\bar{I} \in \mathcal{U}(\gp_1, \cdots, \gp_n)$ for
$C$ is a {\em social model} of $C$ if $ST_{C}(\bar{I})=\bar{I}$.
\end{definition}

Social models are defined as fixpoints of the operator $ST_{C}$. Given a social
interpretation $\bar{I}$, $ST_{C}(\bar{I})$ contains:

\begin{enumerate}
\item for each program $\gp$ in the SOLP collection $C$, the labelled versions
(w.r.t. $\gp$) of the heads of non-tolerance rules, such that the body is true w.r.t.
$\bar{I}$ (According to Definition~\ref{def:SC_sat}, all the SCs included in the body
are checked w.r.t. the given social interpretation $\bar{I}$).

\item for each program $\gp$ in the SOLP collection $C$, the labelled versions
(w.r.t. $\gp$) of the arguments of the predicates $\mathit{okay}$ occurring in the
heads of tolerance rules, such that both the rule body is true w.r.t.~$\bar{I}$ and
the predicate argument is true for $\gp$ w.r.t.~$\bar{I}$.

\end{enumerate}

Observe that the social immediate consequence operator $ST_{C}$ works differently
from the autonomous immediate consequence operator $AT_\gp$ (see
Definition~\ref{def:ATp}), since the former exploits all the programs  -- and the
social conditions included -- of a given SOLP collection $C$, while the latter
operates only within a given program $\gp$, where the social conditions have been
removed.

\begin{definition}\label{def:SOS}
\abbrev{$\mathcal{SOS}(\gp_1,  \cdots, \gp_n)$}{The set of all social models of
$\gp_1, \cdots, \gp_n$} Given a SOLP collection $\{\gp_1, \cdots, \gp_n\}$, the {\em
Social Semantics}
 of $\gp_1,  \cdots, \gp_n$ is the set
$ \mathcal{SOS}(\gp_1,  \cdots, \gp_n) = \{ \bar{M} \mid \bar{M} \in
    \mathcal{U}(\gp_1,  \cdots, \gp_n) \ \And \
    \bar{M} \mbox{ is a social model of } \gp_1,  \cdots, \gp_n \}\mbox{.}
$
\end{definition}

$\mathcal{SOS}(\gp_1,  \cdots, \gp_n)$ is the set of all social models of
$\gp_1, \cdots, \gp_n$.

Now, we introduce an important property holding for social models, i.e. they are {\em
supported} in the associated SOLP collection. The next definition gives the notion of
supportness for a social model.

\begin{definition}\label{def:supportedness}
Given a SOLP collection $C=\{\gp_1, \cdots, \gp_n\}$ and a social model $M \in
\mathcal{SOS}(\gp_1, \cdots, \gp_n)$, $M$ is {\em supported} in $C$ if $\forall \gp
\in C, \forall a \in Var(\gp_1) \ \cup \cdots \ \cup \ Var(\gp_n) ,$ in case $a_{\gp}
\in M,$ then at least one of the following holds:

\begin{itemize}
\item[(1)] $\exists r \mid r \in \gp \ \And \ \mathit{head}(r)=a \ \And \ \mathit{body}(r)$ is true
w.r.t. $M$;

\item[(2)] $\exists r \mid r \in \gp \ \And \ \mathit{head}(r)=\mathit{okay}(a) \
\And \  a$ is true for $\gp$ w.r.t $M$ $\And \ \mathit{body}(r)$ is true w.r.t $M$.
\end{itemize}
\end{definition}

The property is stated in the following theorem.

\begin{theorem}\label{the:SOS-supported}
Given a SOLP collection $C=\{\gp_1, \cdots, \gp_n\}$, $\forall M \in
\mathcal{SOS}(\gp_1,  \cdots, \gp_n)$,  $M$ is supported in $C$.
\end{theorem}

\begin{proof}
By contradiction, assume that $M \in \mathcal{SOS}(\gp_1, \cdots, \gp_n)$ and $M$ is not supported in
$C$. As a consequence,

$$\exists \gp, \exists a \mid \gp \in C \ \And \ a \in \bigcup_{\gp \in C} Var(\gp) \ \And \ a_{\gp} \in M
 \mbox{ and both the following conditions hold:}$$

\begin{description}
\item[](1) $\forall r \in \gp,$ it holds that $\mathit{head}(r)=a \Rightarrow \mathit{body}(r)$ is
false w.r.t. $M$;

\item[](2) $\forall r \in \gp,$ it holds that $\mathit{head}(r)=\mathit{okay}(a)
\Rightarrow a$ is false for $\gp$ w.r.t. $M$ $\And \ \mathit{body}(r)$ is false
w.r.t. $M$.
\end{description}

It is easy to see that, according to Definition~\ref{def:ST_C}
(page~\pageref{def:ST_C}), $a_{\gp} \not\in ST_C(M)$. Now, since, according to
the hypothesis, $a_{\gp} \in M$, it holds that $ST_C(M) \neq M$. Thus $M$ is
not a social model and we have reached a contradiction.
\end{proof}

\begin{example}
Consider the following SOLP collection $C=\{\gp_1, \gp_2\}$:

$
\begin{array}{llr}
    \gp_1:&   a \leftarrow  b, [\gp_2]\{c\} & (r_1)\\
    \gp_2:&   \ \ \ \leftarrow  c & (r_2)\\
\end{array}
$

It holds that $AFP(\gp_1)= \{ \{a,b\},\emptyset\}$, $AFP(\gp_2)= \{ \emptyset \}$.
Thus, there exist two candidate social interpretations, namely
$I_1=\{a_{\gp_1},b_{\gp_1}\}$, $I_2=\emptyset$.

Since both $\mathit{body}(r_1)$ and $\mathit{body}(r_2)$ are false w.r.t. $I_1$, it
holds that $ST_C(I_1)=\emptyset$. As a consequence, $I_1$ is not a social model of
the SOLP collection $C$. Concerning the social interpretation $I_2$ it is easy to see
that $ST_C(I_2)=\emptyset$. Hence, $I_2$ is a social model of the SOLP collection
$C$.

Now, consider a slightly different SOLP collection $C' = \{\gp'_{1}, \gp'_{2}\}$:

$
\begin{array}{llr}
    \gp'_{1}:&   a \leftarrow  b, [\gp'_{2}]\{c\} & (r'_1)\\
    \gp'_{2}:&  c \leftarrow   & (r'_2)\\
\end{array}
$

It holds that: $AFP(\gp'_1)= \{ \{a,b\},\emptyset\}$, $AFP(\gp'_2)= \{ \{c\} \}$.
Thus, we can build the following candidate social interpretations:
$I_1=\{a_{\gp'_1},b_{\gp'_1}, c_{\gp'_2}\}$ and $I_2=\{c_{\gp'_2}\}$.

Now, since $ST_{C'}(I_1)=\{a_{\gp'_1},c_{\gp'_2}\}$ and $\{
a_{\gp'_1},c_{\gp'_2} \} \neq I_1$, $I_1$ is not a social model of the
collection $C'=\{\gp'_1, \gp'_2\}$. Finally, $ST_{C'}(I_2)=I_2$, hence $I_2$ is
a social model of $C'$. It is easy to see that $I_2$ is supported in $C'$.
\end{example}

Now, by means of a complete example, we illustrate the notions introduced above.

\begin{example}

Three agents are represented by the SOLP collection $C=\{P_1, P_2, P_3\}$ next:

\begin{center}
$
{\small
\begin{array}{lrl}
\vspace{0,1cm}
P_1: & go\_party \leftarrow & [2,]\{go\_party, [1,]\{guitar\}\}\\
P_2: & go\_party \leftarrow & [P_3]\{go\_party\}\\
\vspace{0,1cm}
     & guitar \leftarrow & \naf bad\_weather, go\_party\\
P_3: & go\_party \leftarrow & \naf bad\_weather\\
\end{array}
}
$
\end{center}

The intended meaning of the above SOLP programs is the following: agent $P_1$
goes to the party only if there are at least other two agents which go there
and such that at least one of them brings the guitar with him. Agent $P_2$ goes
to the party only if agent $P_3$ goes too. Moreover, in case agent $P_2$ goes
and the weather is not bad, then he thinks it is safe to bring the guitar with
him. Finally, agent $P_3$ goes to the party if there is not evidence of bad
weather.

It is easy to see that $\mathcal{SOS}(\gp_1,  \gp_2, \gp_3) = \{ \bar{I}\}$, where
$\bar{I}$ is the intended model of the collection $C$ and
$\bar{I}=\{go\_party_{P_1}, go\_party_{P_2}, guitar_{P_2}, go\_party_{P_3} \}$.

Indeed, it holds that $AFP(P_1)=\{\{go\_party\}\}$, $AFP(P_2)=\{\{go\_party,
guitar\}\}$ and $AFP(P_3)=\{\{go\_party\}\}$. Now, note that the candidate social
interpretation $\bar{I}$ is a social model of $C$, since it holds that
$ST_C(\bar{I})=\bar{I}$. Finally, it is easy to see that $\bar{I}$ is supported in
$C$.

\end{example}

\section{Translation to Logic Programming with Aggregates}\label{sect:translation}
In this section we give the translation from SOLP under the Social Semantics to logic
programming with aggregates\footnote{We choose the syntax of the non-disjunctive
fragment of $\DLP$~\cite{DLPA03}, denoted as $\LPA$ in the sequel of the
section.\abbrev{$\LPA$}{The non-disjunctive fragment of logic programming with
aggregates supported by the DLV system} The DLV system~\cite{Leone02} can be used to
compute the social models of the SOLP programs.} under the Stable Model Semantics. We
assume that the reader is familiar with the Stable Model Semantics~\cite{GelLif88}.
Given a traditional logic program $\gp$, we denote by $SM(\gp)$ the set of all the
stable models of $\gp$. \abbrev{$SM(\gp)$}{The set of all the stable models of $\gp$}
For the sake of presentation, the most complicated proofs are placed
in~\ref{app:proofs}.

Our goal is the following: given a collection of SOLP programs we have to generate a
single $\LPA$ program whose stable models are in one-to-one correspondence with the
social models of the collection. To this aim we perform the two following
tasks:\label{tasks} ($a$) we generate a $\LPA$ program by means of a suitable
transformation of all the SCs occurring in the SOLP programs of the collection; ($b$)
we obtain another logic program by processing the original SOLP programs in such a
way that the SCs are replaced by suitable atoms. Finally, we merge the two programs
obtained from tasks ($a$) and ($b$) into a single $\LPA$ program. At the end of the
section, we present a comprehensive example (Example~\ref{exa:translation})
describing  the whole translation process.

Next we describe how task ($a$) is performed. The first step is the translation of a
single SC and the extension of such a translation to all the SCs included in a social
rule, a SOLP program and a SOLP collection, respectively. As a result of task ($a$),
a single $\LPA$ program is generated which represents the translation of the social
conditions occurring in the SOLP collection. In order to have fresh literals that
allow us to encode -- in such a program -- the truth of social conditions, we need a
mechanism to generate auxiliary atoms that are in one-to-one correspondence with the
social conditions occurring in a SOLP program.

\begin{definition}\label{def:rho&g}
Given a SOLP program $\gp$, we define $USC^\gp=\bigcup_{r \in \gp}\bigcup_{n \geq
0} MSC^{\langle \gp,r,n \rangle}$.
Moreover, let $L^\rho$ and $L^g$ be two sets of
literals such that both
(1) $Var(\gp)$, $L^\rho$, and $L^g$ are disjoint sets
and (2) $|L^\rho|= |L^g| = |USC^\gp|$.
We define two one-to-one mappings: $\rho:USC^\gp \rightarrow L^\rho$ and
$g:USC^\gp \rightarrow L^g$.

\end{definition}

\abbrev{$USC^\gp$}{The set of all the SCs (at any nesting depth) in $\gp$}
\abbrev{$\rho(s), g(s)$}{Atoms and predicates associated with the social condition
$s$}

Observe that, according to the definition of the set $MSC^{\langle \gp,r,n \rangle}$
(see page~\pageref{def:MSC}), $USC^\gp$ is the set of all the SCs (at any nesting
depth) in $\gp$. Thus, given a SC $s$ included in a SOLP program $\gp$, the mapping
$\rho$ (resp. $g$), returns the auxiliary atom $\rho(s)$ (resp. the predicate $g(s)$)
such that it is fresh, i.e. it does not occur in $\gp$. We will explain next how
$\rho(s)$ and $g(s)$ are exploited by the translation process.

The following definition enables the translation of a single social condition $s$ of
a given program $\gp$ of a SOLP collection $C$. Observe that this definition is
recursive in order to produce the translation of every social condition nested in
$s$. Such a translation produces two sets of rules that we reference as
$GUESS^{\gp}(s)$ and $CHECK^{\gp}(s)$, respectively. Informally, the rules in the set
$GUESS^{\gp}(s)$ aim at verifying properties concerning atoms belonging to other SOLP
programs different from $\gp$. These properties are then checked according to the
selection condition of $s$ (i.e. $cond(s)$) by means of the rules included in the set
$CHECK^{\gp}(s)$.

In the definition, $\rho(s)_{\gp}$ denotes the atom $\rho$ labelled atom w.r.t. $\gp$
(see Definition~\ref{def:labelled_int}) and it is derived in case the social
condition $s$ is true for $\gp$ in $C$ w.r.t. a given social interpretation. With a
little abuse of notation, $(g(s))(x)_{\gp}$ denotes the predicate $g(s)$ labelled
w.r.t. $\gp$, having argument $x$.

\abbrev{$\Psi^{\gp}(s)$}{The translation of a single SC $s$ of a SOLP program $\gp$}

\begin{definition}\label{def:guess&check}
Given a SOLP collection $SP=\{\gp_1, \cdots, \gp_n\}$, an integer $j$ ($1 \leq
j \leq n$), a SOLP program $\gp_j \in SP$ and a social condition $s \in
USC^{\gp_j}$, we define the {\em SC translation of $s$} as the $\LPA$ program
$\Psi^{\gp_j}(s) = GUESS^{\gp_j}(s) \ \cup \ CHECK^{\gp_j}(s)$, where both
$GUESS^{\gp_j}(s) =$
\[
 = \left\{
  \begin{array}{rlr}
   \{ (g(s))(k)_{\gp_j} \leftarrow & \bigwedge_{b \in \mathit{content}(s)} b_{\gp_k}\}, &
   \mbox{ if } \mathit{cond(s)=[\gp_k]} \ \And \ (1 \leq k \leq n),\\

   & & \\
   \{ (g(s))(i)_{\gp_j} \leftarrow & \bigwedge_{b \in \mathit{content}(s)} b_{\gp_i}
      \ \And & \\
   & \multicolumn{2}{l}{\bigwedge_{s' \in skel(s)} \rho(s')_{\gp_j} \mid
      1 \leq i \neq j \leq n \} \ \cup}  \\

   \multicolumn{3}{l}{ \bigcup_{s' \in skel(s)} GUESS^{\gp_j}(s'),
                                        \qquad \qquad \qquad \qquad
                                        \qquad \quad \mbox{ if } \mathit{cond(s)=[l,h]}\mbox{,}}\\

  \end{array}
  \right.
\]
$CHECK^{\gp_j}(s) =$
\[
 = \left\{
  \begin{array}{llr}
   \multicolumn{3}{l}{\{ \rho(s)_{\gp_j} \leftarrow  (g(s))(k)_{\gp_j} \},
   \ \ \quad \qquad \qquad  \mbox{ if } \mathit{cond(s)=[\gp_k]} \ \And \ (1 \leq k \leq n),}\\
   & \\
   \{ \rho(s)_{\gp_j} \leftarrow & l \leq \mathtt{\#count}\{ K:
   (g(s))(K)_{\gp_j}, K \neq j\} \leq h \} \ \cup & \\
   \multicolumn{3}{l}{ \bigcup_{s' \in skel(s)} CHECK^{\gp_j}(s'),
   \qquad \qquad \qquad \qquad \qquad \quad \mbox{if } \mathit{cond(s)=[l,h]}}\\

  \end{array}
  \right.
\]

\noindent and {\tt \#count} denotes an aggregate function which returns the
cardinality of a set of literals satisfying some conditions~\cite{DLPA03}.

\end{definition}

The reader may find an instance of application of the above transformation in the
final example (Example~\ref{exa:translation}, page~\pageref{exa:translation}
). Now,
by means of the next definition, we extend the scope of the above translation to a
social rule, a SOLP program and a SOLP collection.

\begin{definition}\label{def:SCtrans}
Given a SOLP program $\gp$, a social rule $r \in \gp$ and a SOLP collection $\{\gp_1,
\cdots, \gp_n\}$, we define:

\begin{enumerate}

\item the {\em SC translation of r} as the $\LPA$ program $T^{\gp}(r) = \bigcup_{s
\in MSC^\gp} \Psi^{\gp}(s)$;

\item the {\em SC translation of $\gp$} as the $\LPA$ program $W^{\gp} = \bigcup_{r
\in \gp} T^{\gp}(r)$;

\item the {\em SC translation of the collection} as the $\LPA$ program $C(\gp_1,
\cdots, \gp_n) = \bigcup_{1 \leq i \leq n} W^{\gp_i}$.

\end{enumerate}

\end{definition}

\abbrev{$T^{\gp}(r)$}{The translation of the SCs included in a social rule $r$ of
$\gp$}

\abbrev{$W^{\gp}$}{The translation of the SCs included in $\gp$}

\abbrev{$C(\gp_1, \cdots, \gp_n)$}{The $\LPA$ program resulting from the translation
of all the SCs included in the SOLP collection \{$\gp_1, \cdots, \gp_n$\}}

Observe that given a SOLP program $\gp$, for any classical rule $r \in \gp$,
$T^{\gp}(r) = \emptyset$. As a consequence, for any  program $\gp$ with no social
rules, it holds that $W^{\gp} = \emptyset$. $C(\gp_1, \cdots, \gp_n)$ denotes the
$\LPA$ program obtained from the processing of all the SCs included in the SOLP
collection \{$\gp_1, \cdots, \gp_n$\}. The generation of $C(\gp_1, \cdots, \gp_n)$ is
the final step of the task ($a$) within the whole translation machinery.

Now, we describe task ($b$). We introduce a suitable mapping from SOLP programs to
traditional logic programs\footnote{Note that, differently from task ($a$), the logic
program here generated do not contain aggregates.}, and then we apply such a
transformation to each SOLP program in a given SOLP collection. Finally, we combine
the traditional logic programs so obtained into a single program.

Before introducing the mapping, we need a preliminary processing of all tolerance
rules in a SOLP program. This is done by means of the following transformation:

\begin{definition}\label{def:SOLP_hat}
Given a SOLP program $\gp$, we define the SOLP program $\hat{\gp} = \gp \setminus
TR(\gp) \ \cup \  \{p \leftarrow p \ \And \ \mathit{body}(r)
\mid r \in TR(\gp) \ \And \ \mathit{head}(r)=\mathit{okay}(p)\}$.
\end{definition} Note that $\hat{\gp}$ is obtained from $\gp$ by
replacing each tolerance rule $\mathit{okay(p)} \leftarrow \mathit{body}$ with the
rule $p \leftarrow p, \mathit{body}$.

\abbrev{$\hat{\gp}$}{A logic program obtained from $\gp$ after a rewriting of the
tolerance rules occurring in it}

The next step is giving a mapping from a SOLP program to a traditional logic program.

\begin{definition}\label{def:gamma'P}
\abbrev{$\Gamma'()$}{The mapping from SOLP programs to traditional logic programs}
Let $\gp$ be a SOLP program. We define the program $\Gamma'(\hat{\gp})$ over the set
of atoms $Var (\Gamma'(\hat{\gp})) = \{ a_\gp \ | \ a \in Var(A(\hat{\gp})) \} \cup
\{ a'_{\gp} \ | \ a \in Var(A(\hat{\gp})) \} \cup \{ sa_\gp \ | \ a \in
Var(A(\hat{\gp})) \} \cup \{\fail_\gp\}$ as $\Gamma'(\hat{\gp})=S'_1(\hat{\gp}) \cup
S'_2(\hat{\gp}) \cup S'_3(\hat{\gp})$, where $S'_1(\hat{\gp})$, $S'_2(\hat{\gp})$ and
$S'_3(\hat{\gp})$ are defined as follows:

$
\begin{array}{ll}
S'_1(\hat{\gp}) = & \{ a_\gp   \leftarrow  \naf a'_\gp  \ | \ a \in Var(A(\hat{\gp}))
\} \cup
\{ a'_\gp   \leftarrow  \naf a_\gp  \ | \ a \in Var(A(\hat{\gp})) \},\\
& \\
S'_2(\hat{\gp}) = &\{ sa_\gp  \leftarrow  b_\gp^1 , \cdots , b_\gp^n,
\rho(s_1)_\gp, \cdots, \rho(s_m)_\gp \ \mid\\
& \ a \leftarrow b_1, \cdots b_n, s_1, \cdots, s_m \in \gp \},\\
& \\
S'_3(\hat{\gp}) =  & \{ \fail_\gp  \leftarrow  \naf \fail_\gp, sa_\gp , \naf a_\gp \
| \ a \in Var(A(\hat{\gp})) \} \cup\\
&\{ \fail_\gp  \leftarrow  \naf \fail_\gp, a_\gp , \naf sa_\gp \ | \ a \in Var(A(\hat{\gp})) \}\\
\end{array}
$

where $A()$ is the autonomous reduction operator (see page~\pageref{def:A}).

\end{definition}

In words, given a SOLP program $\gp$, first a program $\hat{\gp}$ is produced
(according to Definition~\ref{def:SOLP_hat}) such that all the predicates
$okay()$ occurring in it are suitably translated. Then, according to
Definition~\ref{def:gamma'P}, three sets of standard logic rules are generated
from $\hat{\gp}$, referenced as $S'_1(\hat{\gp})$, $S'_2(\hat{\gp})$ and
$S'_3(\hat{\gp})$. Observe that atoms occurring in these sets are labelled
w.t.r. the source program $\gp$ in order not to generate name mismatch in the
final merging phase. Informally, the set $S'_1(\hat{\gp})$ guesses atoms that
are candidates to be included in a social model. By means of the  rules
included in the set $S'_2(\hat{\gp})$, atoms that are supported by a social
rule are inferred. The atoms denoted by $\rho(s_i)_\gp$ ($1 \leq i \leq m$) are
in one-to-one correspondence with those generated by $W^\gp$ (see
Definition~\ref{def:SCtrans}) and represent the social conditions occurring in
$\gp$. Finally, the set $S'_3(\hat{\gp})$ ensures that an atom is derived by
means of some rule in $S'_2(\hat{\gp})$ iff it is also guessed by some rule in
$S'_1(\hat{\gp})$.

The next definition introduces a logic program representing the translation of the
whole SOLP collection.

\begin{definition}\label{def:P'_u}
Given a SOLP collection $\{\gp_1, \cdots, \gp_n\}$, we define the program
$P'_u = \bigcup_{1 \leq i \leq n} \Gamma'(\hat{\gp_i})$.
\end{definition}

\abbrev{$P'_u$}{The traditional program resulting from the translation of all the
SOLP programs in a SOLP collection, where the SCs are replaced by $\rho$-atoms}
$P'_u$ is obtained by combining the translations of all the SOLP programs in a given
SOLP collection, where the social conditions are replaced by $\rho$-atoms. The
generation of $P'_u$ concludes task ($b$) of the translation process. Then, the
program $P'_u$ is merged with the $\LPA$ program $C(\gp_1, \cdots, \gp_n)$ --
obtained as a result of task ($a$) --  in order to enable the social conditions
(recall that $C(\gp_1, \cdots, \gp_n)$ contains the $\rho$-atoms as heads of rules,
thus allowing the activation of some rule bodies in $P'_u$). Finally, the social
models of the SOLP collection \{$\gp_1, \cdots, \gp_n$\} can be found by computing
the stable models of the logic program $P'_u \cup C(\gp_1, \cdots, \gp_n)$.

Once we have described how the translation mechanism proceeds, we need to demonstrate
that it is sound and complete. To this aim, we have to prove the following results:

\begin{itemize}

\item[(1)] The $\rho$-atoms occurring in $C(\gp_1, \cdots, \gp_n)$ are in one-to-one
correspondence with true SCs for \{$\gp_1, \cdots, \gp_n$\}.

\item[(2)] A one-to-one correspondence exists between the social models of $\gp_1,
\cdots, \gp_n$ and the stable models of the $\LPA$ program $P'_u \cup C(\gp_1,
\cdots, \gp_n)$.

\end{itemize}

First, we prove item (1) above.

\newcommand{\mainlemma}
{ Given a SOLP collection $SP=\{\gp_1, \cdots, \gp_n\}$, an integer $j$ ($1 \leq j
\leq n$), a SOLP program $\gp_j \in SP$, a social interpretation $\bar{I}$ for $SP$
and a SC $s \in MSC^{\gp_j}$, it holds that $s \mbox{ is true for $\gp_j$ w.r.t. }
\bar{I}$ iff $\exists M \in SM(C(\gp_1, \cdots, \gp_n) \cup Q)$ s.t. $\rho(s)_{\gp_j}
\in M$, where $Q=\{a\leftarrow \mid a \in \bar{I}\}$. }

\begin{lemma}\label{lem:SC true iff rho in M}
\mainlemma
\end{lemma}

\begin{proof}
See \ref{app:proofs}.
\end{proof}

Intuitively, a given social interpretation $\bar{I}$ will infer rule heads in
$C(\gp_1, \cdots, \gp_n)$. These are either labelled $\rho$-atoms or labelled
$g$-predicates. Lemma~\ref{lem:SC true iff rho in M} states that the $\rho$-atoms
occurring in $C(\gp_1, \cdots, \gp_n)$ are in one-to-one correspondence with true
social conditions. Now, since those $\rho$-atoms occur also in rule bodies of $P'_u$,
in order to replace
 the corresponding SCs (recall Definitions~\ref{def:gamma'P}
and~\ref{def:P'_u}), they contribute to infer rule heads in $P'_u$, which represent
elements in a social model of the collection \{$\gp_1, \cdots, \gp_n$\}.

Our intention is to compute the social models of $\gp_1, \cdots, \gp_n$ in terms of
the stable models of the logic program $P'_u \cup C(\gp_1, \cdots,
\gp_n)$\footnote{This can be efficiently accomplished by using the DLV
system~\cite{Leone02}}. Thus, we must prove item (2). To this aim, let us recall from
\cite{BucGot02} some definitions and results that we shall use later.

\begin{definition}[\citeANP{BucGot02} \citeyearNP{BucGot02}]\label{def:[]BucGot}
Let $\gp$ be a traditional logic program and $M \subseteq Var(\gp)$. We denote by
$[M]_{\gp}$ the set $\{ a_\gp \ | \ a \in M \} \cup \{ a'_\gp \ | \ a \in Var
(\gp) \setminus M \} \cup \{ sa_\gp \ | \ a \in M \}$.
\end{definition}

\begin{definition}[\citeANP{BucGot02} \citeyearNP{BucGot02}]\label{def:gammaP}
Let $\gp$ be a positive program. We define the program $\Gamma(\gp)$ over the
set of atoms $Var (\Gamma(\gp)) = \{ a_\gp \ | \ a \in Var(\gp) \} \cup \{
a'_\gp \ | \ a \in Var(\gp) \} \cup \{ sa_\gp \ | \ a \in Var(\gp) \} \cup
\{\fail_\gp\}$ as the union of the sets of rules $S_1$, $S_2$ and $S_3$,
defined as follows:
\[
\begin{array}{ll}
S_1 = & \{ a_\gp   \leftarrow  \naf a'_\gp  \ | \ a \in Var(\gp) \}
\cup \{ a'_\gp   \leftarrow  \naf a_\gp  \ | \ a \in Var(\gp) \}\\
\end{array}
\]
\[
\begin{array}{ll}
S_2 = &\{ sa_\gp  \leftarrow  b_\gp^1 , ... , b_\gp^n \ |
\ a \leftarrow b_1, \ldots b_n \in \gp \}\\
\end{array}
\]
\[
\begin{array}{ll}
S_3 =  & \{ \fail_\gp  \leftarrow  \naf \fail_\gp, sa_\gp , \naf a_\gp \ | \
a \in Var(\gp) \} \cup\\
&\{ \fail_\gp  \leftarrow  \naf \fail_\gp, a_\gp , \naf sa_\gp \ | \
a \in Var(\gp) \}\mbox{.}\\
\end{array}
\]
\end{definition}

Note that Definition~\ref{def:gamma'P} (introducing $\Gamma'$) can be viewed as
an extension of the above
definition, since the former takes into account social conditions, while the latter
does not. In fact, the transformations $\Gamma'$ and $\Gamma$ produce the same result
in case of programs with no social conditions.

\begin{proposition}\label{pro:gamma_eq_gamma'}
Given a SOLP program $\gp$, it holds that
$\Gamma'({A(\hat{\gp})})=\Gamma({A(\hat{\gp})})$.
\end{proposition}
\begin{proof}
Since $A(\hat{\gp})$ contains no social condition, it is easy to see that,
according to Definitions~\ref{def:gamma'P} and~\ref{def:gammaP}
(pages~\pageref{def:gamma'P} and~\pageref{def:gammaP}), $S'_1(A(\hat{\gp})) =
S_1(A(\hat{\gp}))$, $S'_2(A(\hat{\gp})) = S_2(A(\hat{\gp}))$ and
$S'_3(A(\hat{\gp})) = S_3(A(\hat{\gp}))$. As a consequence,
$\Gamma'({A(\hat{\gp})})=\Gamma({A(\hat{\gp})})$.
\end{proof}

\begin{lemma}[\citeANP{BucGot02} \citeyearNP{BucGot02}]\label{lem:newequivalence}
Let $\gp$ be a traditional logic program. Then
\[
SM(\Gamma(\gp)) = \bigcup_{F \in FP(\gp)} \{ [F]_\gp \}\mbox{.}
\]
\end{lemma}

Once we have recalled the results from~\cite{BucGot02}, we introduce some more
results that we shall use in order to prove item (2).

\begin{proposition}\label{pro:FP_eq_AFP}
Let $\gp$ be a SOLP program. Then $FP(A(\hat{\gp}))=AFP(\gp)$.
\end{proposition}
\begin{proof}

First observe that, given a SOLP program $\gp$, it holds that $A(\hat{\gp})=
\hat{Q}$ where $Q=A(\gp)$, i.e. the result of the joint application of the two
operators $A()$ and $\hat{\ }$ is invariant w.r.t. to the order of application.
In fact, according to the definitions of both $A()$ (see page~\pageref{def:A})
and $\hat{\ }$ (see Definition~\ref{def:SOLP_hat} on
page~\pageref{def:SOLP_hat}), it is easy to see that the former operates only
on social conditions, while the latter does not, since it operates on both
standard atoms and $\mathit{okay}$ predicates. Thus, the two operators have
disjoint application domains. Hence, the order of application is not relevant.

As a result it holds that, $FP(A(\hat{\gp}))=FP(\hat{Q})$ where $Q=A(\gp)$ and,
according to the traditional definition of fixpoint of a logic program
(page~\pageref{def:FP}), $FP(\hat{Q})=\{X \mid T_{\hat{Q}}(X) = X \ \And \ X \in
2^{Var(\hat{Q})}\}$.

Now, according to the definition of the classical immediate consequence operator
(page~\pageref{def:Tp}), $T_{\hat{Q}}(X)=\{ head(r) \mid r \in \hat{Q} \ \And \
\mathit{body}(r) \mbox{ is true w.r.t. } X\}$, moreover, according to
Definition~\ref{def:SOLP_hat} (page~\pageref{def:SOLP_hat}), $\hat{Q}= Q \setminus
TR(Q) \ \cup \ \{a \leftarrow a, \mathit{body}(r) \mid r \in TR(Q)
 \ \And $ $\ \mathit{head}(r) = \mathit{okay}(a) \}$.
As a consequence, $T_{\hat{Q}}(X)=$

$
\begin{array}{rl}

= & \{ head(r) \mid r \in Q \setminus TR(Q) \ \And \
                       \mathit{body}(r) \mbox{ is true w.r.t. } X\} \ \cup\\
       & \{ a \mid \mathit{head}(r) = \mathit{okay}(a) \ \And \ r \in TR(Q) \ \And \
               (a \ \And \ \mathit{body}(r)) \mbox{ is true w.r.t. } X\}= \\
        = & AT_\gp(X) \mbox{ (see Definition~\ref{def:ATp}, page~\pageref{def:ATp}).}
\end{array}
$

It is easy to see that $FP(A(\hat{\gp})) = FP(\hat{Q}) = \{X \mid T_{\hat{Q}}(X) = X \ \And \ X \in 2^{Var(\hat{Q})}\}=\\
\{X \mid AT_\gp(X) = X \ \And \ X \in 2^{Var(A(\hat{\gp}))}\}$.

Now, observe that $Var(A(\hat{\gp}))=Var^*(A(\gp))$ (see Definition~\ref{def:ATp}),
since after the application of the operator $\hat{\ }$ to $A(\gp)$, each predicate
$\mathit{okay}(p)$ is replaced by its argument $p$, and, according to
Definition~\ref{def:ATp}, for each predicate $okay(p)$ appearing in $A(\gp)$,
$okay(p)$ does not occur in $Var^*(A(\gp))$, but the argument $p$ does.

As a consequence, it holds that $\{X \mid AT_\gp(X) = X \ \And \ X \in 2^{Var(A(\hat{\gp}))}\}=\\
 \{X \mid AT_\gp(X) = X \ \And \ X \in 2^{Var^*(A(\gp))}\} = AFP(\gp)$.

\end{proof}

Now we extend Definition~\ref{def:[]BucGot} and Lemma~\ref{lem:newequivalence}, given
in \cite{BucGot02}, to SOLP programs.

\begin{definition}\label{def:[]}
Let $\gp$ be a (SOLP) program and $M \subseteq Var(\gp)$. We denote by
$[M]_{\gp}$ the set $\{ a_\gp \ | \ a \in M \} \cup \{ a'_\gp \ | \ a \in Var(A(\hat{\gp}))
 \setminus M \} \cup \{ sa_\gp \ | \ a \in M \}$.
\end{definition}

\abbrev{$[\ ]_{\gp}$}{Operator that produces a set of auxiliary atoms labelled w.r.t.
$\gp$ and used in the translation process} Given a (SOLP) program $\gp$, the operator
$[\ ]_{\gp}$ produces a set of auxiliary atoms labelled w.r.t. $\gp$. Those atoms are
used in the translation process. Observe that the above definition extends
Definition~\ref{def:[]BucGot}, since in case $\gp$ is a traditional logic program,
then $ Var(A(\hat{\gp}))= Var(\gp)$ and thus the two definitions match.

The above results are now exploited in order to prove that, by applying the above
operator $[\ ]_{\gp}$  to the autonomous fixpoints of a given SOLP program $\gp$, we
obtain the stable models of the translation of the autonomous version of $\gp$.

\begin{lemma}\label{pro:SM_eq_union}
Given a SOLP program $\gp$, it holds that:
\[
SM(\Gamma'({A(\hat{\gp})})) =  \bigcup_{F \in AFP(\gp)}\{[F]_\gp\}\mbox{.}
\]

\end{lemma}
\begin{proof}
By virtue of Proposition~\ref{pro:gamma_eq_gamma'}
(page~\pageref{pro:gamma_eq_gamma'}), $
\Gamma'({A(\hat{\gp})})=\Gamma({A(\hat{\gp})})$. As a consequence,
$SM(\Gamma'({A(\hat{\gp})}))=SM(\Gamma({A(\hat{\gp})}))$. Now, denoting
$A(\hat{\gp})$ by $Q$,  by virtue of Lemma~\ref{lem:newequivalence}
(page~\pageref{lem:newequivalence}), $SM(\Gamma(Q))= \bigcup_{F \in FP(Q)}
\{[F]_{Q}\}$. According to Definition~\ref{def:[]BucGot}
(page~\pageref{def:[]BucGot}), $[F]_{Q}=\{ a_Q \ | \ a \in F \} \cup \{ a'_Q \
| \ a \in Var(Q) \setminus F\} \cup \{ sa_Q \ | \ a \in F \}$.

Now, recall that $Q=A(\hat{\gp})$. $Q$ represents the SOLP program $P$, after
the application of both operators $\hat{\ }$ and $A()$. As a consequence, atoms
in $[F]_Q$ are labelled w.r.t. $\gp$. Observe that atoms in
$\Gamma'({A(\hat{\gp})})$ are labelled w.r.t. $\gp$ too. Therefore, with a
little abuse of notation, we can write $[F]_{Q}=\{ a_{\gp} \ | \ a \in F \}
\cup \{ a'_{\gp} \ | \ a \in Var(A(\hat{P})) \setminus F\} \cup \{ sa_{\gp} \ |
\ a \in F \}$ $=[F]_{\gp}$, according to Definition~\ref{def:[]}
(page~\pageref{def:[]}) and since $Q$ is a traditional logic program.

Now, we have obtained that $SM(\Gamma({A(\hat{\gp})})) = \bigcup_{F \in FP(A(\hat{\gp}))} \{[F]_{\gp}\}$.

Since,  by virtue of Proposition~\ref{pro:FP_eq_AFP}
(page~\pageref{pro:FP_eq_AFP}), $FP(A(\hat{\gp}))=AFP(\gp)$, it results that

$ \displaystyle{
\bigcup_{F \in FP(A(\hat{\gp}))} \{[F]_{\gp}\} = \bigcup_{F \in AFP(\gp)}\{[F]_\gp\}
}$\mbox{.}
\end{proof}

Now we extend Lemma~\ref{pro:SM_eq_union} to a whole SOLP collection, but first let
us recall the following result from~\cite{EitGot97}.

\begin{lemma}[\citeANP{EitGot97} \citeyearNP{EitGot97}]\label{pro:union}
Let $\gp = \gp_1 \cup \gp_2$ be a program such that $Var(\gp_1) \cap Var(\gp_2) =
\emptyset$.
Then
\[
SM(\gp) = \bigcup_{M_1 \in SM(\gp_1), M_2 \in SM(\gp_2) } \{ M_1 \cup M_2 \}\mbox{.}
\]

\end{lemma}

\begin{lemma}\label{pro:SM(Pu)_eq_union}
Given a SOLP collection $\{\gp_1, \cdots, \gp_n\}$, consider the following sets:

$
\begin{array}{lrl}
(1) & P_u = & \bigcup_{1 \leq i \leq n}\Gamma'(A(\hat{\gp_i})), \\
(2) & SM(P_u), & \mbox{ and} \\
(3) & B = & \left\{ \{(F^1)_{\gp_1} \cup (G^1)_{\gp_1} \} \ \bigcup \cdots \bigcup \
                  \{(F^n)_{\gp_n} \cup (G^n)_{\gp_n} \} \mid  \right. \\
& & \left. \forall i \ 1 \leq i \leq n \ F^i \in AFP(\gp_i) \ \And \ (G^i)_{\gp_i} =
     [F^i]_{\gp_i}
\setminus (F^i)_{\gp_i} \right\}\mbox{.} \\
\end{array}
$

\noindent It holds that $SM(P_u)=B$.
\end{lemma}
\begin{proof} For each $i$ and $j$ such that
$1 \leq i \neq j \leq n$, according to Definition~\ref{def:gamma'P} (page~\pageref{def:gamma'P}),
it holds that $Var
(\Gamma'(A(\hat{\gp_i})))
\cap Var
(\Gamma'(A(\hat{\gp_j}))) = \emptyset$. It is easy to see that:

$
\begin{array}{rl}

SM(P_u) = & SM(\bigcup_{1 \leq i \leq n}\Gamma'(A(\hat{\gp_i})))= \mbox{(by virtue of Lemma~\ref{pro:union}, page~\pageref{pro:union})}\\

   = & \displaystyle{\bigcup_{M^1 \in SM(\Gamma'(A(\hat{\gp_1}))), \cdots, M^n \in SM(\Gamma'(A(\hat{\gp_n})))}
      \{ M^1 \cup \cdots \cup M^n \}}\mbox{.}\\
      \\
\end{array}
$

Note that, for each $i$ ($1 \leq i \leq n$), $M^i \in SM(\Gamma'(A(\hat{\gp_i}))$
and, by virtue of Lemma~\ref{pro:SM_eq_union} (page~\pageref{pro:SM_eq_union}),
$SM(\Gamma'(A(\hat{\gp_i}))=\bigcup_{F \in AFP(\gp)}\{[F]_\gp\}$. Thus,

$
\begin{array}{rl}

  & \displaystyle{\bigcup_{M^1 \in SM(\Gamma'(A(\hat{\gp_1}))), \cdots, M^n \in SM(\Gamma'(A(\hat{\gp_n})))}
       \{ M^1 \cup \cdots \cup M^n \}}=\\

   = & \displaystyle{\bigcup_{F^1 \in AFP(\gp_1), \cdots, F^n \in AFP(\gp_n)}}
      \{ [F^1]_{\gp_1} \cup \cdots \cup [F^n]_{\gp_n} \}\mbox{.}\\

\end{array}
$

Now, for each $i$ ($1 \leq i \leq n$), let us denote by $(G^i)_{\gp_i}$ the set
$[F^i]_{\gp_i} \setminus (F^i)_{\gp_i}$.
It is easy to see that:

$
\begin{array}{rl}

 & \displaystyle{\bigcup_{F^1 \in AFP(\gp_1), \cdots, F^n \in AFP(\gp_n)}}
      \{ [F^1]_{\gp_1} \cup \cdots \cup [F^n]_{\gp_n} \} = \\

 = & \displaystyle{\bigcup_{F^1 \in AFP(\gp_1), \cdots, F^n \in AFP(\gp_n)}}
      \{ \{(F^1)_{\gp_1} \cup (G^1)_{\gp_1} \} \ \cup \cdots \cup \
                  \{(F^n)_{\gp_n} \cup (G^n)_{\gp_n} \} \} = \\
 = & B\mbox{.}\\

\end{array}
$

\end{proof}

Before proving item (2) we need a further definition, introducing the notion of a set
of $\rho$-atoms  and $g$-predicates associated, by virtue of Lemma~\ref{lem:SC true
iff rho in M}, with the social conditions true for a given SOLP program w.r.t. a
social interpretation.

\begin{definition}\label{def:SAT}
Given a SOLP collection $SP=\{\gp_1, \cdots, \gp_n\}$, a social interpretation
$\bar{I}$ for $SP$, a SOLP program $\gp \in SP$ and a SC $s \in MSC^{\gp}$, let
$Q=\{a\leftarrow \mid a \in \bar{I}\}$. We define the set $SAT^{\gp}_{\bar{I}}(s) =$
$$ = \left\{h \mid h=\mathit{head}(r), r \in \Psi^{\gp}(s) \
\And \ (\exists M \in SM(C(\gp_1, \cdots, \gp_n) \cup Q) \mid h \in
M)\right\}\mbox{.}$$
\end{definition}
\abbrev{$SAT$}{A set of $\rho$-atoms  and $g$-predicates associated with the social
conditions true for a given SOLP program w.r.t. a social interpretation}

Observe that in case $s$ is true for $\gp$ w.r.t. $\bar{I}$, $SAT^{\gp}_{\bar{I}}(s)$
includes the atom $\rho(s)_{\gp}$ and those heads of the rules in $\Psi^{\gp}(s)$
(recall from Definition~\ref{def:guess&check} that $\Psi^{\gp}(s)= GUESS^{\gp}(s)
\cup CHECK^{\gp}(s)$) corresponding to both the social condition $s$ and the SCs
nested in $s$.

Finally, we are ready to prove item (2). The next theorem states that a one-to-one
correspondence exists between the social models in $\mathcal{SOS}(\gp_1, \cdots,
\gp_n)$ and the stable models of the $\LPA$ program  $P'_u \cup C(\gp_1, \cdots,
\gp_n)$.

\newcommand{\maintheorem}
{

Given a SOLP collection $SP=\{\gp_1, \cdots, \gp_n\}$, it holds that $A=B$, where:

$
\begin{array}{rlr}
A = & SM(P'_u \cup C(\gp_1, \cdots,\gp_n)) & \mbox{and} \\
B = & \{ \bar{F} \cup \bar{G} \cup \bar{H} \mid
  & \\
 & \bar{F} = \bigcup_{1 \leq i \leq n} (F^i)_{\gp_i} \ \And \
  F^i \in AFP(\gp_i) \ \And \
  \bar{F} \in \mathcal{SOS}(\gp_1, \cdots, \gp_n) \ \And & (1) \\
 & \bar{G} = \bigcup_{1 \leq i \leq n} (G^i)_{\gp_i} \ \And \
   (G^i)_{\gp_i} = [F^i]_{\gp_i} \setminus (F^i)_{\gp_i} \ \And & (2) \\
 &  \bar{H} = \bigcup_{1 \leq i \leq n} (H^i)_{\gp_i} \ \And \
    (H^i)_{\gp_i} = \bigcup_{s \in MSC^{\gp_i}} SAT^{\gp_i}_{\bar{F}}(s) \} \mbox{.} & (3) \\
\end{array}
$

}

\begin{theorem}\label{the:1to1correspndence}

\maintheorem

\end{theorem}

\begin{proof}

See \ref{app:proofs}.

\end{proof}

As a result of the above theorem, each stable model $X$ of the program $P'_u \cup
C(\gp_1, \cdots,\gp_n)$ may be partitioned in three sets: $\bar{F}$ (representing the
corresponding social model of the SOLP collection), $\bar{G}$ and $\bar{H}$ (each
including auxiliary literals needed by the translation). Thus, it is possible to find
the social models of $\gp_1, \cdots, \gp_n$ by a post-processing of the stable models
of $P'_u \cup C(\gp_1, \cdots,\gp_n)$, which drops the sets $\bar{G}$ and $\bar{H}$.

\begin{example}\label{exa:translation}
Before closing the section, we present the following logic program $\gp=P'_u \cup
C(\gp_1, \cdots, \gp_4)$ resulting from the translation of the SOLP collection
$\{\gp_1, \cdots, \gp_4\}$ presented  in Example~\ref{exa:wedding} (see
Section~\ref{sect:introduction}, Table~\ref{tab:wedding example}).

\begin{center}
$ {\small
\begin{array}{rrl}

r_1: & go\_wedding_{P_1} \leftarrow & \naf go\_wedding'_{P_1}\\
r_2: & go\_wedding'_{P_1} \leftarrow & \naf go\_wedding_{P_1}\\
r_3: & \mbox{s}go\_wedding_{P_1} \leftarrow & \rho\_1\_1_{P_1}\\
r_4: & fail_{P_1} \leftarrow & \naf fail_{P_1}, \mbox{s}go\_wedding_{P_1}, \naf go\_wedding_{P_1}\\
r_5: & fail_{P_1} \leftarrow & \naf fail_{P_1}, go\_wedding_{P_1}, \naf \mbox{s}go\_wedding_{P_1}\\
r_6: & go\_wedding_{P_2} \leftarrow & \naf go\_wedding'_{P_2}\\
r_7: & go\_wedding'_{P_2} \leftarrow & \naf go\_wedding_{P_2}\\
r_8: & drive_{P_2} \leftarrow & \naf drive'_{P_2}\\
r_9: & drive'_{P_2} \leftarrow & \naf drive_{P_2}\\
r_{10}: & \mbox{s}go\_wedding_{P_2} \leftarrow & go\_wedding_{P_2}\\
r_{11}: & \mbox{s}drive_{P_2} \leftarrow & drive_{P_2}, go\_wedding_{P_2}\\

\end{array}
}$

$ {\small
\begin{array}{rrl}
r_{12}: & fail_{P_2} \leftarrow & \naf fail_{P_2}, \mbox{s}go\_wedding_{P_2}, \naf go\_wedding_{P_2}\\
r_{13}: & fail_{P_2} \leftarrow & \naf fail_{P_2}, go\_wedding_{P_2}, \naf \mbox{s}go\_wedding_{P_2}\\
r_{14}: & fail_{P_2} \leftarrow & \naf fail_{P_2}, \mbox{s}drive_{P_2}, \naf drive_{P_2}\\
r_{15}: & fail_{P_2} \leftarrow & \naf fail_{P_2}, drive_{P_2}, \naf \mbox{s}drive_{P_2}\\
r_{16}: & go\_wedding_{P_3} \leftarrow & \naf go\_wedding'_{P_3}\\
r_{17}: & go\_wedding'_{P_3} \leftarrow & \naf go\_wedding_{P_3}\\
r_{18}: & \mbox{s}go\_wedding_{P_3} \leftarrow & \rho\_1\_1_{P_3}\\
r_{19}: & fail_{P_3} \leftarrow & \naf fail_{P_3}, \mbox{s}go\_wedding_{P_3}, \naf go\_wedding_{P_3}\\
r_{20}: & fail_{P_3} \leftarrow & \naf fail_{P_3}, go\_wedding_{P_3}, \naf \mbox{s}go\_wedding_{P_3}\\
r_{21}: & g\_1\_1(2)_{P_1} \leftarrow & go\_wedding_{P_2}\\
r_{22}: & g\_1\_1(3)_{P_1} \leftarrow & go\_wedding_{P_3}\\
r_{23}: & \rho\_1\_1_{P_1} \leftarrow &  1 <= \#count\{K : g\_1\_1(K)_{P_1}, K<>1\} <= 3\\
r_{24}: & g\_1\_1(2)_{P_3} \leftarrow & go\_wedding_{P_2}, \naf drive_{P_2}\\
r_{25}: & \rho\_1\_1_{P_3} \leftarrow & g\_1\_1(2)_{P_3}\\

\end{array}
} $
\end{center}
\end{example}

Observe that, according to Definition~\ref{def:gamma'P},
$\Gamma'(\hat{\gp_1})=S'_1(\hat{\gp_1}) \cup S'_2(\hat{\gp_1}) \cup
S'_3(\hat{\gp_1})$, where $S'_1(\hat{\gp_1})=\{r_1, r_2\}$,
$S'_2(\hat{\gp_1})=\{r_3\}$ and $S'_3(\hat{\gp_1})=\{r_4, r_5\}$. Concerning the SOLP
program $\gp_2$, $\Gamma'(\hat{\gp_2})=S'_1(\hat{\gp_2}) \cup S'_2(\hat{\gp_2}) \cup
S'_3(\hat{\gp_2})$, where $S'_1(\hat{\gp_2})=\{r_6, r_7, r_8, r_9\}$,
$S'_2(\hat{\gp_2})=\{r_{10}, r_{11}\}$ and $S'_3(\hat{\gp_2})=\{r_{12}, r_{13},
r_{14}, r_{15}\}$. $\Gamma'(\hat{\gp_3})=S'_1(\hat{\gp_3}) \cup S'_2(\hat{\gp_3})
\cup S'_3(\hat{\gp_3})$, where $S'_1(\hat{\gp_3})=\{r_{16}, r_{17}\}$,
$S'_2(\hat{\gp_3})=\{r_{18}\}$ and $S'_3(\hat{\gp_3})=\{r_{19}, r_{20}\}$. Finally,
$\Gamma'(\hat{\gp_4})=\emptyset$ since $\gp_4$ is empty. Recall that
$P'_u=\Gamma'(\hat{\gp_1}) \cup \Gamma'(\hat{\gp_2}) \cup \Gamma'(\hat{\gp_3}) \cup
\Gamma'(\hat{\gp_4})$ (see Definition~\ref{def:P'_u}).

Now, according to Definition~\ref{def:SCtrans}, $C(\gp_1, \cdots, \gp_4)=\{W^{\gp_1}
\cup W^{\gp_2} \cup W^{\gp_3} \cup W^{\gp_4}\}$, where $W^{\gp_1}=\{r_{21}, r_{22},
r_{23}\}$, $W^{\gp_2}$ and $W^{\gp_4}$ are empty (since $\gp_2$ and $\gp_4$ do not
include any social rule) and, finally, $W^{\gp_3}=\{r_{24}, r_{25}\}$. It is easy to
check that the stable models of $\gp$ correspond, by virtue of
Theorem~\ref{the:1to1correspndence}, to the social models of the SOLP programs
$\gp_1, \cdots, \gp_4$.

\section{Social Models and Joint Fixpoints}\label{sect:social_models_jfp}
\abbrev{COLP}{COmpromise Logic Programming} In this section we show that the social
semantics extends the JFP semantics~\cite{BucGot02}. Basically, COLP programs are
logic programs which also contain {\em tolerance} rules (named {\em okay} rules) that
are rules of the form $okay(p) \leftarrow body(r)$. The semantics of a collection of
COLP programs is defined over traditional programs obtained from the COLP programs by
translating each rule of the form $okay(p) \leftarrow body(r)$ into the rule $p
\leftarrow p, body(r)$. The semantics of a collection $\gp_1,  \cdots, \gp_n$ of COLP
programs is defined in \cite{BucGot02} in terms of joint (i.e., common) fixpoints (of
the {\em immediate consequence operator}) of the logic programs obtained from $\gp_1,
\cdots, \gp_n$ by transforming {\em okay} rules occurring in them (as shown above).

First we need some preliminary definitions and results. We define a translation from
COLP programs~\cite{BucGot02} to SOLP programs:
\begin{definition}\label{def:SOLP_trans}

 Given a COLP program $\gp$ and an integer $n \geq 1$, the {\em SOLP
translation} of $\gp$ is a SOLP program
\[
\sigma^n(\gp) = \{ \sigma^n(r) \mid r \in \gp \},
\]
where
\[
\sigma^n(r)=\left\{
\begin{array}{ll}
  \mathit{head}(r) \leftarrow [n-1, n-1] \mathit{head}(r), \mathit{body}(r)
        &  \mbox{if $r$ is a classical rule,}\\
  \mathit{okay}(p) \leftarrow [n-1, n-1] p, \mathit{body}(r)
        & \mbox{if $r$ is an {\em okay} rule.}\\
\end{array}
\right.
\]

\end{definition}

\abbrev{$\sigma^n(\gp)$}{The translation of a COLP program $\gp$ into a SOLP program}

\begin{definition}\label{def:COLP_hat}

 Given a COLP program $\gp$, let $\mathit{OKAY}(\gp)$ be the set of all the
{\em okay} rules included in $\gp$. We define $\hat{\gp}= \gp \setminus
\mathit{OKAY}(\gp) \cup \{p \leftarrow p, \mathit{body}(r) \mid r \in OKAY(\gp)
\ \And \ \mathit{head}(r) = \mathit{okay}(p)\}$.
\end{definition} \abbrev{$\mathit{OKAY}(\gp)$}{The set of
{\em okay} rules of the COLP program $\gp$} Informally, for any given COLP program
$\gp$, $\hat{\gp}$ is a traditional logic program obtained from $\gp$ by replacing
the head of each {\em okay} rule with the argument of the predicate {\em okay} and
then adding such an argument to the body of the rule.

\begin{lemma}\label{lem:FP_FPa}
Given a COLP program $\gp$, then:
 \[
\forall n \geq 1 \quad FP(\hat{\gp}) = AFP(\sigma^n(\gp))\mbox{.}
\]
\end{lemma} \begin{proof}
First, observe that, according to Definition~\ref{def:ATp}
(page~\pageref{def:ATp}), all SCs occurring in the program
$\sigma^n(\hat{\gp})$ are discarded in order to compute the autonomous
fixpoints. As a consequence, the equivalence holds for any value of the
parameter $n$. Now, it is easy to see that the proof follows directly from
Definitions~\ref{def:auto_fp},~\ref{def:SOLP_trans} and~\ref{def:COLP_hat} (see
page~\pageref{def:auto_fp},~\pageref{def:SOLP_trans}
and~\pageref{def:COLP_hat}, respectively).
\end{proof}

\begin{lemma}\label{lem:SOS_jointFP}
Let $\gp_1,  \cdots, \gp_n$ be a set of COLP programs and $Q_1, \cdots, Q_n$ be
SOLP programs such that $Q_i = \sigma^n(\gp_i)$. Then:
\[
\mathcal{SOS}(Q_1, \cdots, Q_n) = \{ (F^1)_{Q_1} \cup \cdots \cup (F^n)_{Q_n} \mid
\forall i,j \quad 1 \leq i \neq j \leq n, \quad F^i = F^j \}\mbox{.}
\]
\end{lemma} \begin{proof} By contradiction let us assume that
\[
 \mbox{(1)  } \exists \bar{M} \mid \bar{M} = (F^1)_{Q_1} \cup \cdots \cup (F^n)_{Q_n}
       \ \And \ \bar{M} \in \mathcal{SOS}(Q_1, \cdots, Q_n), \mbox{ and}
\]
\[
\mbox{(2)  } \exists i,j \quad 1 \leq i \neq j \leq n \mid F^i \neq F^j.
\]
Thus, without loss of generality there exists $h \in F^i$ s.t. $h_{Q_i} \in
(F^i)_{Q_i}$ and $h_{Q_j} \not\in (F^j)_{Q_j}$. As a consequence, $h_{Q_i} \in
\bar{M}$ and $h_{Q_j} \not\in \bar{M}$. Now, we have reached a contradiction because,
according to Definitions~\ref{def:SOS} and~\ref{def:SOLP_trans}
(pages~\pageref{def:SOS} and~\pageref{def:SOS}), $h_{Q_i} \in \bar{M}$ only if for
each $k$, ($1 \leq k \neq i \leq n$), $h_{Q_k} \in \bar{M}$, thus $\bar{M} \not\in
\mathcal{SOS}(Q_1, \cdots, Q_n)$ (contradiction).

\end{proof}

The next theorem states that the JFP semantics is a special case of the social
semantics. $JFP(\gp_1,  \cdots, \gp_n)$ denotes the set of the joint fixpoints of the
collection of COLP programs $\gp_1,  \cdots, \gp_n$. \abbrev{$JFP(\gp_1,  \cdots,
\gp_n)$}{The set of the joint fixpoints of the COLP programs $\gp_1, \cdots, \gp_n$}

\begin{theorem}\label{the:SOLP_generalizes_COLP}
Let $\gp_1,  \cdots, \gp_n$ ($n \geq 1$) be COLP programs and $C=\{Q_1, \cdots, Q_n\}$
be a collection of SOLP programs such that $Q_i = \sigma^n(\gp_i)$. Then:
\[
\mathcal{SOS}(Q_1, \cdots, Q_n) = \left\{ \bigcup_{1 \leq i \leq n} (F)_{Q_i} \mid F
\in JFP(\gp_1,  \cdots, \gp_n) \right\}\mbox{.}
\]
\end{theorem} \begin{proof} {($\supseteq$).} First, we show that
\[
\forall F \in JFP(\gp_1,  \cdots, \gp_n), \quad  \bigcup_{1 \leq i \leq n} (F)_{Q_i}
\in \mathcal{SOS}(Q_1, \cdots, Q_n)\mbox{.}
\]

By contradiction, let us assume that $\exists \bar{M} = \bigcup_{1 \leq i \leq n}
(F)_{Q_i} \mid \bar{M} \not\in \mathcal{SOS}(Q_1, \cdots, Q_n)$.

Thus, there exists an integer $j, \ 1 \leq j \leq n$, such that either:
\begin{description}
\item[] (1) $F \not\in AFP(Q_j)$, or

\item[] (2) $F \in AFP(Q_j) \ \And \ ST_C(\bar{M}) \neq \bar{M}$.
\end{description}
In case (1), by virtue of Lemma~\ref{lem:FP_FPa} (page~\pageref{lem:FP_FPa}),
$F \not\in FP(\hat{P_j})$ which contradicts the hypothesis that $F \in
JFP(\gp_1, \cdots, \gp_n)$.

In case (2), it holds that either:
\begin{description}

\item[] ($a$) $\exists h, Q_j \mid Q_j \in C \ \And \ h_{Q_j} \in \bar{M} \ \And \
              h_{Q_j} \not\in ST_C(\bar{M})$, or

\item[] ($b$) $\exists h, Q_j \mid Q_j \in C \ \And \ h_{Q_j} \not\in \bar{M} \ \And \
              h_{Q_j} \in ST_C(\bar{M})$.

\end{description}

If condition ($a$) occurs, then, for each rule $r \in Q_j,$ s.t.
$\mathit{head}(r) = h$ or $\mathit{head}(r)=\mathit{okay}(h)$, it results that
$\mathit{body}(r)$ is false w.r.t. $\bar{M}$, because $h_{Q_j} \not\in
ST_C(\bar{M}$ (recall Definition~\ref{def:ST_C}, page~\pageref{def:ST_C}).

Now, since $h \in F$ and $F \in AFP(Q_j)$, according to
Definition~\ref{def:SOLP_trans} (page~\pageref{def:SOLP_trans}), it holds that
for each rule $r \in Q_j,$ s.t. $\mathit{head}(r) = h$ or
$\mathit{head}(r)=\mathit{okay}(h)$, the SC $[n-1, n-1]h$ (introduced by the
transformation $\sigma$)
 is false for $Q_j$ w.r.t. $\bar{M}$.

Thus, according to Definition~\ref{def:SC_sat} (page~\pageref{def:SC_sat}),
there exists $k$ ($1 \leq k \neq j \leq n$) s.t. $h_{Q_k} \not\in \bar{M}$.
Now, we have obtained that $h_{Q_j} \in \bar{M}$ and $h_{Q_k} \not\in \bar{M}$.
Since $\bar{M} = \bigcup_{1 \leq i \leq n} (F)_{Q_i}$ and $F \in JFP(\gp_1,
\cdots, \gp_n)$, by virtue of Lemma~\ref{lem:SOS_jointFP}
(page~\pageref{def:SOLP_trans}), we have reached a contradiction. This
concludes the proof in case (2), when condition (a) holds.

Consider, now, that condition ($b$) is true in case (2). According to
Definition~\ref{def:ST_C} (page~\pageref{def:ST_C}), there exists $r \in Q_j$
s.t. $\mathit{head}(r) = h$ or $\mathit{head}(r)=\mathit{okay}(h)$ and
$\mathit{body}(r)$ is true
 w.r.t. $\bar{M}$. As a consequence and according to Definition~\ref{def:SOLP_trans} (page~\pageref{def:SOLP_trans}), $[n-1, n-1]h$ is
true for $Q_j$ w.r.t. $\bar{M}$. Now, according to Definition~\ref{def:SC_sat}
(page~\pageref{def:SC_sat}), for each $k$ ($1 \leq k \neq j \leq n$), $h_{Q_k}
\in \bar{M}$ and $h_{Q_j} \not\in \bar{M}$. Since $\bar{M} = \bigcup_{1 \leq i
\leq n} (F)_{Q_i}$ and $F \in JFP(\gp_1,  \cdots, \gp_n)$, by virtue of
Lemma~\ref{lem:SOS_jointFP} (page~\pageref{lem:SOS_jointFP}), we have reached a
contradiction. This concludes the proof in case (2).

{($\subseteq$).} Now we show that $\forall \bar{M} \in \mathcal{SOS}(Q_1, \cdots, Q_n)$, it holds that
both:

$
\begin{array}{cr}
(1) & \bar{M}=\{\bigcup_{1 \leq i \leq n} (F^i)_{Q_i} \mid \forall i,j \ (1 \leq i \neq j
\leq n), F^i=F^j \} \mbox{ and}\\
(2) & \forall i \ (1 \leq i \leq n), \  F^i \in JFP(\gp_1, \cdots, \gp_n)\mbox{.}

\end{array}
$

First observe that, condition (1) follows directly from
Lemma~\ref{lem:SOS_jointFP} (page~\pageref{lem:SOS_jointFP}).

Now we prove that condition (2) is true. Assume, by contradiction, that $F
\not\in JFP(\gp_1, \cdots, \gp_n)$, where $F=F^1=F^2=\cdots=F^n$ (thanks to
condition (1)). As a consequence, there exists $j$ ($1 \leq j \leq n$) s.t.
$F\not\in FP(\hat{\gp_j})$ (recall that ) $\gp_j$ is a COLP program. Now, by
virtue of Lemma~\ref{lem:FP_FPa} (page~\pageref{lem:FP_FPa}), $F \not\in
AFP(Q_j)$, where $Q_j=\sigma^n(\gp_j)$. Thus, according to
Definition~\ref{def:SOS} (page~\pageref{def:SOS}), $F_{Q_j} \not\subseteq
\bar{M}$, because $F_{Q_j}$ is not an autonomous fixpoint of $Q_j$. This result
contradicts condition (1), which is true,
stating that $F_{Q_j} \subseteq \bar{M}$.
\end{proof}

\section{Complexity Results}\label{sect:complexity}
In this section we introduce some relevant decision problems with respect to the
Social Semantics and discuss their complexity. The analysis is done in case of
positive programs. The extension to the general case is straightforward.

First, we consider the problem of social model existence for a collection of SOLP programs.\\

\noindent PROBLEM $SOS_n$ (social model existence):
\begin{description}
\item[Instance:] A SOLP collection $C=\{\gp_1,\cdots,\gp_n\}$. \item[Question:] Is
$\mathcal{SOS}(\gp_1,\cdots, \gp_n)\neq \emptyset$, i.e., do the programs
$\gp_1,\ldots, \gp_n$ have any social model?
\end{description}
\abbrev{$SOS_n$}{The decision problem ``social model existence''}

\begin{theorem}\label{the:SOS_NP-complete}{\em
The problem $SOS_n$ is NP-complete. }\end{theorem} \begin{proof} {\em (1. Membership)}.\quad In order
to verify that a set of positive SOLP programs admits a social model, it suffices to guess a
candidate social interpretation $\bar{I}$ for $C$ and then to check that $ST_C(\bar{I})=\bar{I}$.
Since the latter task is feasible in polynomial time, then the problem $SOS_n$ is in NP.

{ \em (2. Hardness)}.\quad Observe that the problem $SOS_n$ generalizes the
problem $\JFP$~\cite{BucGot02}, which has been proved to NP-complete. Indeed,
in Definition~\ref{def:SOLP_trans} (page~\pageref{def:SOLP_trans}) a
polynomial-time reduction from $\JFP$ to $SOS_n$, i.e. $\sigma^n(\gp)$, has
been introduced. Moreover, Theorem~\ref{the:SOLP_generalizes_COLP}
(page~\pageref{the:SOLP_generalizes_COLP}) states that any instance of the
problem $\JFP$ can be reduced to an equivalent instance of $SOS_n$, i.e. on
those instances the both problems have the same answers. Thus, we have proven
that the problem $SOS_n$ is NP-hard.
Then the problem $SOS_n$ is NP-complete.
\end{proof}

Indeed, the case of non positive programs is straightforward: since it is NP-complete
to determine whether a {\em single} non-positive program has a fixpoint, it is easy
to see that the same holds for non-positive SOLP programs and autonomous fixpoints.
Thus, checking whether a SOLP collection containing at least one non-positive SOLP
program has a social model is trivially NP-hard. Moreover,
since this problem is easily seen to be in NP, it is NP-complete.\\

Now, we introduce several computationally interesting decision problems associated
with the social semantics. Each of them corresponds to a computational task involving
labeled atom search inside the social models of a SOLP collection.

The traditional approach used for classical non-monotonic semantics of
logic programs, typically addresses the two following problems:
\begin{description}
    \item[]($a$) {\em Skeptical Reasoning}, i.e. deciding whether an atom $x$ occurs in all the models
     of a given program $\gp$;
    \item[]($b$) {\em Credulous Reasoning}, i.e. deciding whether an atom $x$ occurs in some model of
     a given program $\gp$.
\end{description}

Since a social model is a social interpretation, i.e. a set of labeled
atoms, we have to extend the above problems ($a$) and ($b$) by introducing a further search
dimension, expressing the {\em sociality} degree of the agents represented by the
SOLP collection. More informally, we are also interested in {\em how many} SOLP
programs a given atom -- occurring as a labeled atom in a social model -- is
entailed by.

As a consequence, given a collection $C=\{\gp_1,\cdots,\gp_n\}$ of SOLP
programs, a social model $\bar{M}$ of $\gp_1,\cdots,\gp_n$ and an atom $x$, we
distinguish two cases. Either:

\begin{description}
    \item[] (1) for each $\gp$ in $C$,
     $\forall i$ $(1 \leq i \leq n)$, $x_{\gp_i} \in \bar{M}$, or
    \item[] (2) for some $\gp$ in $C$,
     $\exists i \mid 1 \leq i \leq n \ \And \ x_{\gp_i} \in \bar{M}$.
\end{description}

In words, in case (1) the agents corresponding to the SOLP collection $C$ exhibit a greater sociality
degree -- since all of them choose the atom $x$ inside the social model $\bar{M}$ -- than in case (2),
where at least one agent is required to choose $x$ and thus we observe a more individual agent
behaviour.

By combining the problems (a) and (b) with the traditional reasoning tasks (1) and (2),
we obtain the following four decision problems relevant to the social semantics:

\begin{description}

\item[1.] PROBLEM $SS$-$SOS_n$ (socially skeptical reasoning):

\item[Instance:] A SOLP collection $C=\{\gp_1,\cdots,\gp_n\}$ and an atom $x$.

\item[Question:] Does it hold that, for each $\bar{M} \in \mathcal{SOS}(\gp_1, \cdots, \gp_n)$,
$\{x_{\gp_1}, \cdots, x_{\gp_n}\} \subseteq \bar{M}$, i.e. $\forall i$ ($1 \leq i \leq n), x_{\gp_i}
\in \bar{M}$?
\end{description}
\abbrev{$SS$-$SOS_n$}{The decision problem ``socially skeptical reasoning''}

In case the answer to such a problem is positive, then it holds that all the agents always (i.e., in
each social model) choose $x$, since $x_\gp$ occurs in $\bar{M}$,
for each social model $\bar{M}$ and for each SOLP program $\gp$. For instance, this
kind of reasoning could be applied by the government of a given country in order to
know if all citizens, modelled as a collection of SOLP programs, pay taxes.

\begin{description}

\item[2.] PROBLEM $IS$-$SOS_n$ (individually skeptical reasoning):

\item[Instance:] A SOLP collection $C=\{\gp_1,\cdots,\gp_n\}$ and an atom $x$.

\item[Question:] Does it hold that, for each $\bar{M} \in \mathcal{SOS}(\gp_1, \cdots, \gp_n)$,
$\exists i \mid$ 1~$\leq$~$i$~$\leq$~$n$ $\ \And \ x_{\gp_i} \in \bar{M}$?
\end{description}
\abbrev{$IS$-$SOS_n$}{The decision problem ``individually skeptical reasoning''}

In case the answer to such a problem is positive, then it holds that always (i.e., in every social
model) there is at least an agent choosing $x$, since $x_\gp$ occurs in $\bar{M}$,
for each social model $\bar{M}$ and for some SOLP program $\gp$. This kind of
reasoning is useful, for instance, to test if a given action, represented by
$x$, is always performed by at least one agent, no matter the agent is.
For example, consider a family (modelled as a SOLP collection) sharing a car.
The above kind of reasoning could be used in order to check
whether someone gets gasoline each time the car is used.

\begin{description}

\item[3.] PROBLEM $SC$-$SOS_n$ (socially credulous reasoning):

\item[Instance:] A SOLP collection $C=\{\gp_1,\cdots,\gp_n\}$ and an atom $x$.

\item[Question:] Does it hold that, there exists $\bar{M} \in \mathcal{SOS}(\gp_1, \cdots, \gp_n)$,
such that for each $i$ $(1 \leq i \leq n)$, $x_{\gp_i} \in \bar{M}$, i.e. $\{x_{\gp_1}, \cdots,
x_{\gp_n}\} \subseteq \bar{M}$?
\end{description}
\abbrev{$SC$-$SOS_n$}{The decision problem ``socially credulous reasoning''}

In case the answer to such a problem is positive, then at least one social
model exists
whereas all the agents choose $x$, since $x_\gp$ occurs in $\bar{M}$,
for some social model $\bar{M}$ and for each SOLP program $\gp$. As a
consequence, a common agreement on $x$ by the agents may be reached at least in
one case (i.e. in one social model). For instance, this kind of reasoning could
be applied in order to check whether some chance exists that the European
Council of Ministers (modelled as a SOLP collection) unanimously accepts a
country as a new member of the European Union.

\begin{description}

\item[4.] PROBLEM $IC$-$SOS_n$ (individually credulous reasoning):

\item[Instance:] A SOLP collection $C=\{\gp_1,\cdots,\gp_n\}$ and an atom $x$.

\item[Question:] Does it hold that, there exist $\bar{M}$ and $j$ such that (i) $\bar{M} \in
\mathcal{SOS}(\gp_1, \cdots, \gp_n)$ and (ii) it holds that $1 \leq j \leq n \ \And \ x_{\gp_j} \in
\bar{M}$?
\end{description}
\abbrev{$IC$-$SOS_n$}{The decision problem ``individually credulous reasoning''}

In case the answer to such a problem is positive, then it holds that at least one social model exists
whereas at least one agent chooses $x$, since $x_\gp$ occurs in $\bar{M}$,
for some social model $\bar{M}$ and for some SOLP program $\gp$. In such a
case, although there is no common agreement on $x$ by the agents, it holds that
$x$ is chosen by some of them,  at least once (i.e. in one social model). For
example, the above reasoning could be used by a company in order to check whether
a given product is never bought by a group of potential customers (represented
by SOLP programs).\\

The computational complexity of the above problems  is stated by the following
theorems.

\begin{theorem}\label{the:SS-SOS_coNP-complete}{\em
The problem $SS$-$SOS_n$ is coNP-complete. }\end{theorem} \begin{proof} It
suffices to prove that the complementary problem of $SS$-$SOS_n$ is
NP-complete. Such a problem may be described as follows:

\begin{description}
\item[Instance:] A SOLP collection $C=\{\gp_1,\cdots,\gp_n\}$ and an atom $x$.

\item[Question:] Does it hold that there exist $\bar{M}$ and $j$ such that (i) $\bar{M} \in
\mathcal{SOS}(\gp_1, \cdots, \gp_n)$ and (ii) $1 \leq j \leq n \ \And \ x_{\gp_j} \not\in \bar{M}$?
\end{description}

{\em (1. Membership)}.\quad We need to guess a candidate social interpretation
$\bar{M}$ for $C$ and, then, to verify that:

\begin{description}

\item[](i) $\bar{M} \in \mathcal{SOS}(\gp_1, \cdots, \gp_n)$, and

\item[](ii) $\exists j \mid 1 \leq j \leq n \ \And \ x_{\gp_j} \not\in \bar{M}$.

\end{description}

Verifying the above items is feasible in polynomial time. Thus, the complementary problem of
$SS$-$SOS_n$ is in NP.

{ \em (2. Hardness)}.\quad  Now we prove that a reduction from the NP-complete
problem $SOS_n$  to the complementary problem of $SS$-$SOS_n$ is feasible in
polynomial time. Consider an atom $x$ such that, for each $i$ ($1 \leq i \leq
n$), $x \not\in Var(\gp_i)$. It is easy to see that $\mathcal{SOS}(\gp_1,
\cdots, \gp_n) \neq \emptyset $ iff there exists $\bar{M} \in
\mathcal{SOS}(\gp_1, \cdots, \gp_n)$ such that $\{x_{\gp_1}, \cdots,$ $
x_{\gp_n} \} \not\subseteq \bar{M}$. Moreover, in case $\{x_{\gp_1}, \cdots,$ $
x_{\gp_n} \} \not\subseteq \bar{M}$, it results that  $\exists j \mid 1 \leq j
\leq n \ \And \ x_{\gp_j} \not\in \bar{M}$. Thus, $SOS_n$ is polynomially
reducible to the complementary problem of $SS$-$SOS_n$.
\end{proof}

\begin{theorem}\label{the:IS-SOS_coNP-complete}{\em
The problem $IS$-$SOS_n$ is coNP-complete. }\end{theorem}
\begin{proof} It suffices to prove that the complementary
problem of $IS$-$SOS_n$ is NP-complete. Such a problem may be described as follows:

\begin{description}
\item[Instance:] A SOLP collection $C=\{\gp_1,\cdots,\gp_n\}$ and an atom $x$.

\item[Question:] Does it hold that there exists $\bar{M} \in \mathcal{SOS}(\gp_1, \cdots, \gp_n)$ such
that  $\forall j $ $(1 \leq j \leq n), x_{\gp_j} \not\in \bar{M}$, i.e. $\{x_{\gp_1}, \cdots,
x_{\gp_n} \} \not\subseteq \bar{M}$?
\end{description}

{\em (1. Membership)}.\quad We need to guess a candidate social interpretation
$\bar{M}$ for $C$ and, then, to verify that:

\begin{description}

\item[](i) $\bar{M} \in \mathcal{SOS}(\gp_1, \cdots, \gp_n)$, and

\item[](ii) $\{x_{\gp_1}, \cdots, x_{\gp_n} \} \not\subseteq \bar{M}$.

\end{description}

Verifying the above items is feasible in polynomial time. Thus, the complementary problem of
$IS$-$SOS_n$ is in NP.

{ \em (2. Hardness)}.\quad Now we prove that a reduction from the NP-complete
problem $SOS_n$ to the complementary problem of $IS$-$SOS_n$ is feasible in
polynomial time. Consider an atom $x$ such that, for each $i$ ($1 \leq i \leq
n$), $x \not\in Var(\gp_i)$. It is easy to see that $\mathcal{SOS}(\gp_1,
\cdots, \gp_n) \neq \emptyset $ iff there exists $\bar{M} \in
\mathcal{SOS}(\gp_1, \cdots, \gp_n)$ such that $\{x_{\gp_1}, \cdots,$ $
x_{\gp_n} \} \not\subseteq \bar{M}$. Thus, $SOS_n$ is polynomially reducible to
the complementary problem of $IS$-$SOS_n$.
\end{proof}

\begin{theorem}\label{the:SC-SOS_NP-complete}{\em
The problem $SC$-$SOS_n$ is NP-complete. }\end{theorem}
\begin{proof} {\em (1. Membership)}.\quad We need to guess a
candidate social interpretation $\bar{M}$ for $C$ and, then, to verify that:

\begin{description}

\item[](i) $\bar{M} \in \mathcal{SOS}(\gp_1, \cdots, \gp_n)$, and

\item[](ii) $\{x_{\gp_1}, \cdots, x_{\gp_n} \} \subseteq \bar{M}$.

\end{description}

Verifying the above items is feasible in polynomial time. Thus, the problem $SC$-$SOS_n$ is in NP.

{ \em (2. Hardness)}.\quad Now we prove that a reduction from the NP-complete problem
$SOS_n$ exists and it is feasible in polynomial time. consider the SOLP collection
$C'=\{\tau(\gp_1), \cdots, \tau(\gp_n)\}$, where, for each $i$ ($1 \leq i \leq n$),
$\tau(\gp_i)$ is a SOLP program obtained from $\gp_i$ as follows:
$$ \tau(\gp_i) = \gp_i \cup \{x \leftarrow\}\mbox{,} $$
where it holds that, for each $i$ ($1 \leq i \leq n$), $ x \not\in Var(\gp_i)$.
It is easy to see that $\mathcal{SOS}(\gp_1, \cdots, \gp_n)\neq \emptyset$ iff
for each $\bar{M} \in \mathcal{SOS}(\gp_1, \cdots, \gp_n)$,
there exists $\bar{M}' \in \mathcal{SOS}(\tau(\gp_1), \cdots, \tau(\gp_n))$
such that $\bar{M}' = \bar{M} \cup \{x_{\gp_1}, \cdots,$ $ x_{\gp_n} \}$.
Thus, $SOS_n$ is polynomially reducible to $SC$-$SOS_n$.
\end{proof}

\begin{theorem}\label{the:IC-SOS_NP-complete}{\em
The problem $IC$-$SOS_n$ is NP-complete. }\end{theorem}
\begin{proof} {\em (1.
Membership)}.\quad We need to guess a candidate social interpretation $\bar{M}$ for
$C$ and, then, to verify that:

\begin{description}

\item[](i) $\bar{M} \in \mathcal{SOS}(\gp_1, \cdots, \gp_n)$, and

\item[](ii) $\exists j \mid 1 \leq j \leq n \ \And \ x_{\gp_j} \in \bar{M}$.

\end{description}

Verifying the above items is feasible in polynomial time. Thus, the problem $IC$-$SOS_n$ is in NP.

{ \em (2. Hardness)}.\quad Now we prove that a reduction from the NP-complete problem
$SOS_n$ exists and it is feasible in polynomial time. consider the SOLP collection
$C'=\{\tau(\gp_1), \cdots, \tau(\gp_n)\}$, where, for each $i$ ($1 \leq i \leq n$),
$\tau(\gp_i)$ is a SOLP program obtained from $\gp_i$ as follows:
$$ \tau(\gp_i) = \gp_i \cup \{x \leftarrow\}\mbox{,} $$
where it holds that, for each $i$ ($1 \leq i \leq n$), $ x \not\in Var(\gp_i)$.
It is easy to see that $\mathcal{SOS}(\gp_1, \cdots, \gp_n)\neq \emptyset$ iff
for each $\bar{M} \in \mathcal{SOS}(\gp_1, \cdots, \gp_n)$,
there exists $\bar{M}' \in \mathcal{SOS}(\tau(\gp_1), \cdots, \tau(\gp_n))$
such that $\bar{M}' = \bar{M} \cup \{x_{\gp_1}, \cdots,$ $ x_{\gp_n} \}$.
Moreover, since $\{x_{\gp_1}, \cdots,$ $ x_{\gp_n} \} \subseteq \bar{M}'$,
there exists $j$ ($1 \leq j \leq n$) such that $x_{\gp_j} \in \bar{M}'$.
Thus, $SOS_n$ is polynomially reducible to $IC$-$SOS_n$.
\end{proof}

\section{Knowledge Representation with SOLP programs}\label{sect:kr}
In this section, we provide interesting examples showing the capability of our
language of representing common knowledge.

\begin{example}[Seating] We must arrange a seating for a number $n$ of agents (representing,
for instance,  people invited to the wedding party introduced in
Example~\ref{exa:wedding}), with $m$ tables and a maximum of $c$ chairs per
table. Agents who like each other should sit at the same table; agents who
dislike each other should not sit at the same table. Moreover, an agent can
express some requirements w.r.t. the number and the identity of other agents
sitting at the same table. Assume that the $I$-th agent is represented by a
predicate $agent(I)$ and his knowledge base is enclosed in a single SOLP
program. Each program will include both a set of common rules encoding the
problem and the agent's own requirements. The predicate $like(A)$ (resp.
$dislike(A)$) means that the agent A is desired (resp. not tolerated) at the
same table. $table(T)$ represents a table ($1 \leq T \leq m$) and $at(T)$
expresses the desire to sit at table $T$. For instance, the program $\gp_1$
(which is associated with the agent 1) could be written as follows:

\begin{center}
{\small
 $
\begin{array}{lrl}
r_1: & agent(1) \leftarrow & \\
r_2: & \leftarrow & at(T1), at(T2), T1 <> T2 \\
r_3: & at(T) \leftarrow & [c,]\{at(T), agent(P)\}, like(P), table(T) \\
r_4: & \leftarrow & at(T), [1,]\{at(T), agent(P)\}, dislike(P) \\
r_5: & \leftarrow & like(P), dislike(P)\\
r_6: & like(2) \leftarrow & \\
r_7: & dislike(3) \leftarrow & \\
r_8: & okay(like(4)) \leftarrow & \\
r_9: & \leftarrow & at(T), [3,]\{at(T)\} \\
\end{array}
$ }
\end{center}

\noindent where the rules from $r_1$-$r_5$ are common to all the programs (of
course, the argument of the predicate $agent()$ in $r_1$ is suited to the
enclosing program) and the rules $r_6$-$r_9$ express the agent's own
requirements. In detail, the rule $r_2$ states that any agent cannot be seated
at more than one table, the rule $r_3$ means that agent 1 sits at a particular
table $T$ if at least $c$ agents he likes are seating at that table ($c$ is a
given constant). The rule $r_4$ states that it is forbidden that agent 1 shares
a table with at least one or more agents he dislikes.

The rule $r_5$ provides consistency for the predicates $like$ and $dislike$, while
examples of such predicates are reported in rules $r_6$ and $r_7$. The rule $r_8$ is
used to declare that agent 1 tolerates agent 4, i.e. agent 4 possibly shares a table
with agent 1, and finally the rule $r_9$ means that the agent 1 does not want to
share a table with 3 agents or more. Observe that while the rule $r_3$ generates
possible seating arrangements, the rules $r_2$, $r_4$ and $r_9$ discard those which
are not allowed.

\end{example}

\begin{example}[Room arrangement] Consider a house having $m$ rooms. We have to
distribute some objects (i.e. furniture and appliances) over the rooms in such a way that we do not
exceed the maximum number of objects, say $c$, allowed per room. Constraints about the color and/or
the type of objects sharing the same room can be introduced. We assume that each object is represented
by a single program encoding both the properties and the constraints we want to meet. Consider the
following program:

\begin{center}
{\small
$
\begin{array}{lrl}
r_1: & name(cupboard) \leftarrow & \\
r_2: & type(furniture) \leftarrow & \\
r_3: & color(yellow) \leftarrow & \\
r_4: & \leftarrow & at(R1), at(R2), R1 <> R2 \\
r_5: & at(R) \leftarrow & [,2]\{at(R),type(appliance),color(yellow),\\
& & \ \ \ \ \  \ \ [1,1]\{name(fridge)\}\}, room(R)\\
r_6: & at(R) \leftarrow & [,c-3]\{at(R), type(furniture)\}, room(R)\\
r_7: & \leftarrow & at(R), [1,]\{at(R), color(green)\}\\
\end{array}
$ }
\end{center}
\noindent where the properties of the current object are encoded as predicates representing the $name$
(fridge, cupboard, table, \ldots), the $type$ (furniture or appliance), the $color$ and so on (see
rules $r_1$-$r_3$). In particular, the rule $r_4$ states that an object may not be in more than one
room, while by means of the rule $r_5$, we allow no more than two yellow appliances to share the room
with the cupboard, provided that one of them is a fridge. The rule $r_6$ means that we want the
cupboard to be in the same room with any other pieces of furniture, but no more than $c-3$, where $c$
(representing the maximum number of objects per room) is given. Finally, the rule $r_7$ states that
the cupboard cannot share the room with any green object.
\end{example}

\begin{example}[FPGA Design]\label{exa:fpga}
In this example, we represent an extended version of a well-known problem belonging
to the setting of FPGA (Field Programmable Gate Arrays) design, namely {\em
placement}. FPGAs are generic, programmable digital devices providing, in a single
system, a way for digital designers to access thousands or millions of logic gates
arranged in multilevel structures, referred to as {\em modules}, and to program them
as desired by the end user. Placement consists in determining the module positions
within the design area according to given constraints.

Consider a team of $n$ electronic engineers, jointly working on a common FPGA design.
Each of them is responsible for placing a given number of modules inside the chip
design area, which is represented by a square grid of cells. In particular, each
designer must meet a number of constraints concerning either (resp. both) his own
modules or (resp. and) the modules of other designers. Moreover, the total chip area
which is occupied by the modules must either match a given value or be less than a
given value.

This setting can be encoded by a collection of SOLP programs\footnote{Observe
that, according to the implementation of the language that relies on DLV
\cite{Leone02,DLPA03}, individual programs of our SOLP collection adopt the
syntax of DLV, allowing both built-in predicates and standard aggregate
functions.}  $C=\{Q_1, \cdots, Q_n,$ $ P\}$, where for each $i$ ($1 \leq i \leq
n$), $Q_i$ describes designer $i$'s requirements, and $P$ represents an agent
aimed to find admissible solutions to the placement problem. Such solutions are
included into the social models of the SOLP collection $C$. In the following
paragraphs we describe such programs in detail.

We assume that each module is rectangular-shaped and it is described by a predicate
$mod(M,X,$ $Y)$, where $M$ is an identifier and $X$ (resp. $Y$) is the horizontal
(resp. vertical) module size measured in grid cells. First, we describe the program
$Q_i$ (corresponding to rules $r_1$-$r_{28}$), encoding designer $i$'s placement
constraints.

We distinguish among {\em hard} and {\em soft} constraints, respectively. Hard
constraints are common to each program $Q_i$ ($1 \leq i \leq n$) and describe
the placement problem. Soft constraints represent both requirements of a single
designer on his own module's properties and requirements on the properties of
modules owned by another designer.

{\em Hard Constraints.} The following rules encode the placement problem:

\begin{center}
$
{\small
\begin{array}{lrl}

r_1: & grid(n,n) \leftarrow &\\

r_2: & \leftarrow & place(B,X,Y), unplaced(B,X,Y)\\

r_3: & place(B,X,Y) \leftarrow & \naf unplaced(B,X,Y), mod(B,\_,\_), \#int(X),\\
     & & \#int(Y), X<XG, Y<YG, grid(XG,YG)\\

r_4: & unplaced(B,X,Y) \leftarrow & \naf place(B,X,Y), mod(B,\_,\_), \#int(X),\\
     & & \#int(Y), X<XG, Y<YG, grid(XG,YG)\\

r_5: & \leftarrow & place(B,X,Y), mod(B,XB,YB), grid(XG,YG),\\
     & & XT>XG, +(X,XB,XT)\\

r_6: & \leftarrow & place(B,X,Y), mod(B,XB,YB), grid(XG,YG),\\
     & & YT>YG, +(Y,YB,YT)\\

r_7: & \leftarrow & \#count\{B: place(B,\_,\_), mod(B,\_,\_), B=B1\}=0,\\
     & & mod(B1,\_,\_)\\

r_8: & \leftarrow & place(B,X,Y), place(B,X1,\mbox{{\em Y}1}), X<>X1\\
r_9: & \leftarrow & place(B,X,Y), place(B,X1,\mbox{{\em Y}1}), Y<>\mbox{{\em Y}1}\\

r_{10}: & \leftarrow & cell(B,X,Y), cell(B1,X,Y), B<>B1\\

r_{11}: & cell(B,X,Y) \leftarrow & mod(B,XB,YB), place(B,X1,\mbox{{\em Y}1}),  \#int(X),\\
        & & \#int(Y), X1 \leq X,  X<SX, +(X1,XB,SX),\\
 & & \mbox{{\em Y}1} \leq Y, Y<SY, +(\mbox{{\em Y}1},YB,SY)\\

r_{12}: &  \leftarrow & cell(B,X,Y), [1,]\{cell(B1,X1,\mbox{{\em Y}1})\}, X=X1, Y=\mbox{{\em Y}1}\\
\end{array}
}
$
\end{center}

First, the grid sizes are declared (rule $r_1$) and then, after candidate
module positions are guessed (rules $r_2$-$r_4$), several requirements are
checked: (i) a module cannot be placed outside the chip design area (rules
$r_5, r_6$); (ii) all modules must be placed (rule $r_7$); ($iii$) each
module can not be placed more than once (rules $r_8, r_9$); ($iv$) modules
owned by the same designer cannot overlap (rule $r_{10}$). By means of rule
$r_{11}$ the predicate $cell(B,X,Y)$
 is true if module $B$ covers the grid cell at coordinates $X,Y$. Finally, the intended
meaning of the social rule $r_{12}$  is to avoid overlapping of modules owned by
different designers.

{\em Soft Constraints.} The following rules describe examples of constraints that
designer $i$ can specify on his own module's properties, i.e. setting either the
absolute module position (rule $r_{13}$) or that relative to other modules (rules
$r_{14}$-$r_{17}$). For instance, rules $r_{14}$ and $r_{15}$ specify that both
modules 1 and 3 must be placed ($a$) on the same row (represented by the coordinate
Y), and ($b$) such that module 1 is on the left of module 3. Finally, rules $r_{16}$
and $r_{17}$ require that module 1 is placed either 0 or 1 cell far from module 3.
Note that the predicate $place(B,X,Y)$ sets the upper-left corner coordinates of
module $B$ to $X,Y$.

\begin{center}
$
{\small
\begin{array}{lrl}

r_{13}: & place(1,3,4) \leftarrow &\\
r_{14}: & \leftarrow & place(1,X,Y), place(3,X1,\mbox{{\em Y}1}), Y<>\mbox{{\em Y}1}\\
r_{15}: & \leftarrow & place(1,X,Y), place(3,X1,\mbox{{\em Y}1}), X \geq X1\\
r_{16}: & \leftarrow & place(1,X,Y), place(3,X1,\mbox{{\em Y}1}), \#int(X),\\
        & & mod(1,XB,YB), +(X,XM,X1), +(XB,D,XM), D<0\\
r_{17}: & \leftarrow & place(1,X,Y), place(3,X1,\mbox{{\em Y}1}), \#int(X),\\
        &  &  mod(1,XB,YB), +(X,XM,X1), +(XB,D,XM), D>1\\

\end{array}
}
$
\end{center}

In addition, it is possible to encode, by means of social rules, the dependence
of designer $i$'s module properties from those of other designers. For
instance, given an integer $d$, by means of the following rules designer $i$
requires that module 4 is placed on the same row (rule $r_{18}$) as designer
$j$'s module 1 and such that a distance of exactly $d$ cells exists between
them (rules $r_{19},r_{20}$).

\begin{center}
$
{\small
\begin{array}{lrl}
r_{18}: &  \leftarrow & place(4,X4,Y4), [Q_j]\{place(1,X1,\mbox{{\em Y}1})\}, Y4<>\mbox{{\em Y}1}.\\
r_{19}: & \leftarrow & place(4,X4,Y4), [Q_j]\{place(1,X1,\mbox{{\em Y}1})\}, D<>d, +(X4,XM,X1),\\
        & & X4 \leq X1, +(XB4,D,XM), mod(4,XB4,YB4)\\
r_{20}: & \leftarrow & place(4,X4,Y4), [Q_j]\{place(1,X1,\mbox{{\em Y}1})\}, D<>d, +(X1,XM,X4),\\
        & & X1 \leq X4, +(XB1,D,XM), mod(1,XB1,YB1)\\
\end{array}
}
$
\end{center}

In order to ensure that all modules are properly spaced, rules
$r_{21}$-$r_{24}$ (resp. rules $r_{25}$-$r_{28}$) require that modules owned by
designer $i$ (resp. owned by designers $i$ and $j$ such that $j \neq i$) are
mutually spaced by at least $k$ cells, where $k$ is a given integer constant.

\begin{center}
$
{\small
\begin{array}{lrl}
r_{21}:  & \leftarrow &  cell(B,X,Y), cell(B1,X1,\mbox{{\em Y}1}), B<>B1, \#int(D), +(X,D,X1), D<k\\
r_{22}:  & \leftarrow &  cell(B,X,Y), cell(B1,X1,\mbox{{\em Y}1}), B<>B1, \#int(D), +(X1,D,X), D<k\\
r_{23}:  & \leftarrow &  cell(B,X,Y), cell(B1,X1,\mbox{{\em Y}1}), B<>B1, \#int(D), +(Y,D,\mbox{{\em Y}1}), D<k\\
r_{24}:  & \leftarrow &  cell(B,X,Y), cell(B1,X1,\mbox{{\em Y}1}), B<>B1, \#int(D), +(\mbox{{\em Y}1},D,Y), D<k\\
r_{25}:  & \leftarrow &  cell(B,X,Y), [1, ]\{cell(B1,X1,\mbox{{\em Y}1})\}, \#int(D), +(X,D,X1), D<k\\
r_{26}:  & \leftarrow &  cell(B,X,Y), [1, ]\{cell(B1,X1,\mbox{{\em Y}1})\}, \#int(D), +(X1,D,X), D<k\\
r_{27}:  & \leftarrow & cell(B,X,Y), [1, ]\{cell(B1,X1,\mbox{{\em Y}1})\}, \#int(D), +(Y,D,\mbox{{\em Y}1}), D<k\\
r_{28}:  & \leftarrow & cell(B,X,Y), [1, ]\{cell(B1,X1,\mbox{{\em Y}1})\}, \#int(D), +(\mbox{{\em Y}1},D,Y), D<k\\
\end{array}
}
$
\end{center}

Now we describe the SOLP program $P$ (rules $r_{29}$-$r_{38}$), representing an agent
which collects from the designers admissible solutions to the placement problem.
Moreover, by means of additional rules ($r_{39}$-$r_{40}$), $P$ possibly requires
that the placement layout area either is less than or matches a given value.

\begin{center}
$
{\small
\begin{array}{lrl}
r_{29}: &  pcell(X,Y) \leftarrow & [1, ]\{cell(\_,X1,\mbox{{\em Y}1})\}, X=X1, Y=\mbox{{\em Y}1}\\
r_{30}: &  exists\_x\_lt(W) \leftarrow & pcell(W,\_), pcell(W1,\_), W1<W\\
r_{31}: &  lowest\_x(X) \leftarrow &  pcell(X,\_), \ \naf exists\_x\_lt(X)\\
r_{32}: &  exists\_y\_lt(W) \leftarrow & pcell(\_,W), pcell(\_,W1), W1<W\\
r_{33}: &  lowest\_y(Y) \leftarrow & pcell(\_,Y), \ \naf exists\_y\_lt(Y)\\
r_{34}: &  exists\_x\_ht(W) \leftarrow & pcell(W,\_), pcell(W1,\_), W1>W\\
r_{35}: &  highest\_x(X) \leftarrow & pcell(\_,X,\_), \ \naf exists\_x\_ht(X)\\
r_{36}: &  exists\_y\_ht(W) \leftarrow & pcell(\_,W), pcell(\_,W1), W1>W\\
r_{37}: &  highest\_y(Y) \leftarrow &  pcell(\_,Y), \ \naf exists\_y\_ht(Y)\\
r_{38}: &  design\_area(A) \leftarrow &  *(B,H,A), +(X1,B,X2),
        +(\mbox{{\em Y}1},H,\mbox{{\em Y}2}), lowest\_x(X1),\\
& &  highest\_x(X2), lowest\_y(\mbox{{\em Y}1}), highest\_y(\mbox{{\em Y}2})\\

\end{array}
}
$
\end{center}

Social rule $r_{29}$ collects admissible solutions to the placement problem. The
rules from $r_{30}$ to $r_{37}$ are used to represent the smallest rectangle
enclosing all the placed modules. Then, the actual design area is computed by rule
$r_{38}$.

In case an an upper bound $b$ to be satisfied (resp. an exact value $s$ to be
matched) is given, then the following rule $r_{39}$ (resp. $r_{40}$) may be added:

\begin{center}
$
{\small
\begin{array}{llrl}
& r_{39}: &  \leftarrow & design\_area(A), A > b\\
(\mbox{resp.} & r_{40}: &  \leftarrow & design\_area(A), A <> s)\\
\end{array}
}
$
\end{center}

\end{example}

\begin{example}[Contextual Reasoning]
It is interesting to observe that SOLP programs can represent a form of Contextual
Reasoning~\cite{ghidini-giunchiglia_f:2001a}.

Although many definitions of the notion of {\em context} exist in the Artificial Intelligence
literature~\cite{mccarthy_j1:1993b,ghidini-giunchiglia_f:2001a,brezillon99}, we can informally say
that a context is an environment (i.e. a set of facts and the logic rules to perform inference with)
in which the reasoning takes place.

In particular, in~\cite{ghidini-giunchiglia_f:2001a} two key principles of contextual reasoning are
stated: {\em locality} (the reasoning task uses only a subset of the total knowledge available) and
{\em compatibility} (additional constraints among different contexts may be specified to declare those
which are mutually compatible).

Under this perspective, we are interested in representing this feature of commonsense
reasoning, that is, given a problem to be solved, (i) bounding the reasoning to the
knowledge which is strictly needed, the so-called {\em context} of the problem, and
(ii), in case the original context is not suitable to reach a solution, enabling
the use of new information provided by other contexts.

It is interesting to note that SOLP programs are well-suited to represent contexts,
since each of them enables reasoning which takes place both locally, i.e. at the
level of the program knowledge base, and at the level of the other programs'
knowledge bases. Thus, it is easy to model reasoning which involves several different
contexts.

For instance, consider a SOLP program $P_0$ (representing the context of the agent
$A_0$) including a social rule of the form

\begin{center}
{
$ action \leftarrow [P_1] b_1, \cdots, [P_n] b_n $ }
\end{center}
meaning that the agent $A_0$ infers the term {\em action} if each term $b_i$ ($1 \leq i \leq n$) is
inferred by the corresponding agent $A_i$, i.e. $b_i$ is part of an autonomous
fixpoint of $P_i$. This way, the notion of locality is realized by representing each
different context in a separate SOLP program, and the compatibility principle is
pursued by suitably using SCs where member selection conditions identify contexts.

As an example, consider the well-known ``Three Wise Men Puzzle'', first introduced in
~\cite{konoligephd}.

\begin{quote} {\em
A king wishes to determine which of his three wise men is the wisest. He arranges them in a circle so
that they can see and hear each other and tells them that he will put a white or a black spot on each
of their foreheads but that at least one spot will be white. He then repeatedly asks them, ``Do you
know the colour of your spot?''. What do they answer?}
\end{quote}

We represent by means of SOLP programs a slightly simpler version of the puzzle,
where only two wise men are involved:

\begin{center}
$
{\small
\begin{array}{lrl}

r_{1}: &  color(white) \leftarrow & \\
r_{2}: &  color(black) \leftarrow & \\

\hline

\multicolumn{3}{c}{\mbox{\bf King (agent $K$)}}\\

r_{3}: &  wise\_man(1\mbox{..}2) \leftarrow & \\
r_{4}: &  put\_spot(A, black) \leftarrow & \naf put\_spot(A, white), wise\_man(A)\\
r_{5}: &  put\_spot(A, white) \leftarrow & \naf put\_spot(A, black), wise\_man(A)\\
r_{6}: &  \leftarrow & \#count\{S: put\_spot(S, black)\} = N,\\
& & \#count\{A: wise\_man(A)\} = N, \#int(N)\\

r_{7}: & ask\_question(1) \leftarrow &\\

r_{8}: & ask\_question(T2) \leftarrow & ask\_question(T1), \#succ(T1,T2), \#int(T1), \#int(T2)\\
& & \naf [1,]\{answer(white,T1), answer(black,T1)\},\\
\hline

\end{array}
} $

$
{\small
\begin{array}{lrl}

\multicolumn{3}{c}{\mbox{\bf Wise man 1 (agent $W1$)}}\\

r_{9}: & forehead(C) \leftarrow & [K]\{put\_spot(1,C)\}, color(C)\\

r_{10}: & answer(white,1) \leftarrow & [K]\{ask\_question(1)\}, [W2]\{forehead(black)\}\\

r_{11}: & answer(black,1) \leftarrow & [K]\{ask\_question(1)\}, [W2]\{forehead(white),\\
& & answer(white,1)\}\\

r_{12}: & answer(white,2) \leftarrow & [K]\{ask\_question(2)\}, [W2]\{forehead(white),\\
& & \naf answer(white,1), \ \naf answer(black,1)\}\\
\end{array}
} $
\end{center}

Rules $r_1$-$r_2$ are common to each SOLP program. Such rules set the admissible spot
colors. Rules $r_3$-$r_8$ represent the king. Rule $r_3$ sets the number of wise men
(two in this case). By means of rules $r_4$-$r_5$, the king non-deterministically
puts a spot on each wise man's forehead. Rule $r_6$ represents the king's statement
``At least one spot is white''. The king asks the question for the first time (rule
$r_7$) and, after he has asked the question, if no agent gives an answer, then he
asks the question again (rule $r_8$). Rule $r_9$ is used to store into the predicate
$forehead$ the information about the color of the corresponding wise man's spot.
Observe that since each wise man cannot look at his own forehead, then the predicate
$forehead$ is further referenced in SCs only. Rules $r_{10}-r_{11}$ represent the
case of exactly one white spot: after the first time the king asks the question, if
the first wise man sees a black spot on the other wise man's forehead, then he
concludes that his spot is white (rule $r_{10}$). Otherwise, if the second wise man
both has a white spot on his forehead and he answers ``white'', then the first wise
man can conclude that the other wise man has seen a black spot on his forehead. Thus,
the first wise man answers ``black''. Finally, rule $r_{12}$ represents the case of
two white spots. In such a case, after the first question, no wise man can conclude
anything about the color of his own spot. After the second time the king asks the
question, each wise man can answer ``white'' in case he sees a white spot and the
other wise man has not answered the previous king's question. The correctness of such
a statement can be proved by contradiction: if a wise man had a black spot on his
forehead, then the other wise man would have seen it and, thus, he also would have
answered ``white'' after the first king's question.

The SOLP program representing the second wise man (program $W2$) is  easily obtained
from  $W1$ by exchanging the role of the two wise men, i.e. by replacing each
occurrence of the program identifier $W2$ by $W1$.

For the sake of the simplicity we have considered a simple scenario, i.e. two wise
men. It is possible to extend the reasoning encoded in the above programs, in order
to write a general program for $n$ wise men, by exploiting the nesting feature of the
social conditions in such a way that reasoning on both the content and the temporal
sequence of the wise men's statements is enabled.

\end{example}

\section{Related Work}\label{sect:relwork}
\paragraph{Contextual Reasoning} - As pointed out in Section~\ref{sect:kr}, a relationship
 exists between our work
and~\cite{ghidini-giunchiglia_f:2001a}, where the {\em Local Model Semantics} (LMS)
is proposed to reason about contexts.

A survey covering the use of contexts in many fields of Artificial Intelligence can
be found in~\cite{brezillon99}. An approach concerning contextual reasoning
and agent-based systems can be found in ~\cite{climavii-saeger}.

In~\cite{mccarthy_j1:1993b}, the author discusses the notion of context in Artificial
Intelligence in order to solve the problem of {\em generality}, that is every logic
theory is valid within the bounds of a definite context and it is possible to design
a more general context where such a theory is not valid anymore.

Other approaches have been proposed in \cite{buvac-mason_m:1993a} and
\cite{ghidini-giunchiglia_f:2001a}. Moreover, these two works are compared
in~\cite{DBLP:journals/ai/SerafiniB04}. In the former work, the authors
introduce the {\em Propositional Logic of Context}, a modal logic aiming at
formalizing McCarthy's ideas.

An approach which is more closely related to ours is proposed
in~\cite{ghidini-giunchiglia_f:2001a}, where the authors consider a set of
logic languages, each representing a different context, and a suitable
semantics is used to select among sets of {\em local} models, i.e. models
pertaining to a single language, those which satisfy a given {\em
compatibility} condition. Moreover, a proof-theoretical framework for
contextual reasoning, called {\em Multi-Context Systems}, is introduced where
the notions of locality and compatibility are respectively captured by {\em
inference} rules, whose scope is the single language, and {\em bridge} rules
establishing relationships among different languages.

Observe that, in the previous section, we have shown that such a machinery can
be represented by social rules where the body includes only
member-selection-condition-based SCs, each corresponding to a different context
to be included into the reasoning task. As a consequence, we argue that social
rules are more general than bridge rules, since the former provide also
(possibly nested) cardinal-selection-condition-based SCs. Since our work is not
aimed to reason on contexts, a direct comparison with
~\cite{ghidini-giunchiglia_f:2001a} cannot be done, although some
correspondences may be found between the model-theoretical formalizations of
both the Local Model Semantics and the Social Semantics. In particular, we feel
that the latter could be easily adapted to fully enable contextual reasoning
inside logic programming. This is left for future work.

\paragraph{Logic-based Multi-Agent Systems} - A related approach, where the
semantics of a collection of abductive logic agents
is given in terms of the stability of their interaction
can be found in~\cite{DBLP:conf/dalt/BraccialiMST04} where the authors define
the semantics of a multi-agent system via a definition of {\em stability} on
the set of all actions performed by all agents in the system, possibly arising
from their communication and interaction via observation. According to the
authors, a set of actions committed by different agents is {\em stable} if,
assuming that an ``oracle'' could feed each of the agents with all the actions
in the set performed by the other agents, each agent would do exactly what is
in the set. We believe that such a machinery is similar to our fixpoint-based
approach since we guess candidate social interpretations and select those which
are compatible with the SCs of the SOLP collection.

In our work we focused on the formalization of the semantics, assuming a
perfect communication among the agents in such a way that each agent is able to
know the mental state of the others. Using a different approach from ours,
in~\cite{DBLP:conf/atal/SatohY02}, in order to face the possible incompleteness
of information due to communication failures or delays in a multi-agent system,
a default hypothesis is used as a tentative answer and the computation
continues until a reply is received which contradicts with the default.

Another interesting work is the MINERVA agent architecture~\cite{LAP02}, based on
dynamic logic programming~\cite{AlferesLPPP00}. MINERVA is a modular architecture,
where every agent is composed of specialized sub-agents that execute special tasks,
e.g., reactivity, planning, scheduling, belief revision, action execution. A common
internal knowledge base, represented as one or more Multi-dimensional Dynamic Logic
Programs (MDLP)~\cite{LeiteAPPP02}, is concurrently manipulated by its specialized
sub-agents. The MDLPs may encode object-level knowledge, or knowledge about goals,
plans, intentions, etc.

The DALI project~\cite{dalijelia02} is a complete multi-agent platform entirely
written in Prolog. A DALI program results in an agent which is capable of reactive
and proactive behaviour, triggered by several kinds of events. The semantics of a
DALI program is defined in terms of another program, where reactive and proactive
rules are reinterpreted as standard Horn Clause rules.

Laima~\cite{laimadalt05} agents are represented as Ordered Choice Logic
Programs (OCLP)~\cite{oclpagp03} for modelling their knowledge and reasoning
capabilities. Communication between the agents is regulated by uni-directional
channels transporting information based on their answer sets.

IMPACT~\cite{subrahmanian00heterogeneous} is an agent platform where programs may be
used to specify what an agent is either obliged to do, may do, or cannot do on the
basis of deontic operators of permission, obligation and prohibition. IMPACT is
grounded on a solid semantic framework based on the concept of feasible status set,
which describes a set of actions dictated by an agent program that is consistent with
the obligations and restrictions on the agent itself. Agent programs define integrity
constraints, which must be satisfied in order to provide a feasible status set. The
adoption of a logic programming based formalism, and the use of integrity constraints
to define a feasible status set, guarantee agents to behave in a way that some
desired properties hold.

Societies Of ComputeeS (SOCS)~\cite{socs04} is a project that was funded by European
Union. The idea is to provide a computational logic model for the description,
analysis and verification of global and open societies of heterogeneous {\em
computees}. The computee model is proposed as a full-fledged agent model, based on
extended logic programming, allowing to define and study properties that can be
enforced by its operational model.

Since our approach relies on the general notion of social behaviour, it is of course
interesting to illustrate how this concept is dealt with in the related field of
intelligent agents, in order to make evident that -- even in such context -- this
notion takes an important role. Indeed, beside {\em autonomy}, intelligent
agents~\cite{WoolJen95,Wool00} may be required to have {\em social ability}. The
meaning of this concept is two-fold: (1) the presence of a common language for
communication, and (2) the capability of reasoning on the content of communication
acts. Concerning item (1), KQML~\cite{KQML95},
and FIPA ACL~\cite{oai:CiteSeerPSU:478447}, both based on the {\em speech act
theory}~\cite{CL87a}, represent the main efforts done in the last years. The
state-of-the-art literature on item (2) is represented
by~\cite{Wool00,hoek03towards,MMS03}. Social ability allows thus the agent
individuals to have {\em beliefs}, {\em desires} and {\em intentions}
(BDI)~\cite{RaoGeo95,Rao96} as a result of both the mutual communication and the
consequent individual reasoning.

\section{Conclusions and Future Work}\label{sect:conclusions}
In this paper we have proposed a new language, {\em Social Logic Programming} (SOLP),
which extends Compromise Logic Programming and enables social behaviour among a
community of individuals whose reasoning is represented by logic programs.
A rich set of examples shows that the language has very nice capabilities
of representing such a kind of knowledge.
Moreover, we have given a translation from SOLP to logic programming with aggregates and discussed
the computational complexity of several decision problems related to the social
semantics.

Basically, the present paper gives the theoretical core for a multi-agent oriented
software environment, including suitable specialized features, like information
hiding, speech-act mechanisms, security and so on. However, these issues are
interesting directions of our future work.

For instance, information hiding can be implemented as follows. Given a SOLP
collection $C$, we make the following assumption: by default each agent cannot see
into other agents' mind, that is all atoms in each SOLP program are viewed only by
the program itself, i.e. they are private. In order to make some atom public we could
add a suffix, say $P$, to such an atom, i.e. $a$ is meant as private, while $aP$ is
meant as the public version of $a$. Then, by means of a suitable modification of our
translation machinery, any social condition can be activated only on public atoms.

An agent communication machinery can be conceived that relies on the above feature.
In case an agent wants to send a message to another agent, then the former could make
public a suitable set of atoms in such a way that they are visible only to the latter.
This approach could be easily extended to the scenario where one agent wants to send
a message to either a group of agents or to the whole community.

Another feature we intend to include into future extensions of SOLP is the
representation of evolving agent mental states. We believe that a collection of SOLP
programs could be easily managed by some existing logic framework which is tailored
to program update or belief revision tasks, such as, for instance, Dynamic Logic
Programming (DLP)~\cite{AlferesLPPP00}. The resulting system should work in a cyclic
fashion: (i) social models of the SOLP collection are computed, (ii) by exploiting
DLP, the SOLP programs within the collection are possibly updated according to the
intended evolution of agent beliefs and intentions, (iii) the cycle restarts.
Starting from the basic approach proposed in Section~\ref{sect:kr}
(Example~\ref{exa:fpga}), another interesting issue to investigate is the capability
of the language, in the general case, of representing cooperative approaches to
solving combinatorial optimization problems, possibly by introducing some suitable
extensions.

Finally, we plan to enhance the language of SOLP by adding both classical negation
and rule-head disjunction. We expect the former to be easy to implement, while the
integration of latter requires some preliminary study, as the introduction of
disjunction in logic programming always results in a growth of the language
expressivity towards higher levels in the computational complexity hierarchy.

\bibliography{solp}

\appendix

\section{}\label{app:proofs}

In order to improve the overall readability of the paper, a number of lemma and
theorem proofs have been moved to this section.

\begin{thm*}{Lemma}{lem:SC true iff rho in M}
\mainlemma
\end{thm*}

\begin{proof}
Before starting with the proof, let us denote the set of all the SCs occurring in $s$
(plus the SC $s$ itself) by $N(s)$, and the $\LPA$ program $C(\gp_1, \cdots, \gp_n)$
by $\bar{C}$.

($\Rightarrow$). We have to prove that if $s$ is true for $\gp_j$ w.r.t. $\bar{I}$,
then there exists a stable model $M$ of the logic program $\bar{C} \cup Q$
 s.t. the atom $\rho(s)_{\gp_j}$, corresponding to $s$ by means of the translation,
is included in $M$. We proceed by induction on the maximum nesting depth (see
page~\ref{def:depth})
 of the SCs in $N(s)$,
$d=max_{s' \in N(s)}\{depth(s')\}$ $(d \geq 0)$.

({\em Basis}). In case $d=0$, then $N(s)=\{s\}$ and $depth(s)=0$ (recall the
definition of the function depth on page~\ref{def:depth}). Since $|N(s)|=1$, $s$ is a
simple SC, i.e. $skel(s)=\emptyset$. Now, assume by contradiction that $s$ is true
for $\gp_j$ w.r.t. $\bar{I}$ and that
 for each $M \in SM(\bar{C} \cup Q),$ $\rho(s)_{\gp_j} \not\in M$.
 Observe now that it may occur either the following cases:
 (1) $cond(s)=[\gp_k] (1 \leq k \leq n)$,
or (2) $cond(s)=[l,h]$.

In case (1), according to Definition~\ref{def:guess&check}
(page~\pageref{def:guess&check}), there exists a set $S=\{r_1, r_2\} \subseteq
\bar{C} \cup Q$ such that:

\begin{center}
$ {\small
\begin{array}{lrl}

r_1: & \rho(s)_{\gp_j} \leftarrow & (g(s))(k)_{\gp_j}\\
r_2: & (g(s))(k)_{\gp_j} \leftarrow & \bigwedge_{b \in \mathit{content}(s)} b_{\gp_k}
     \\

\end{array}
} $
\end{center}

\noindent and for each $r \in (\bar{C} \cup Q)\setminus S$, $head(r) \neq r_1 \ \And
\ head(r) \neq r_2$.

Since we have assumed that $\rho(s)_{\gp_j} \not\in M$, it is easy to see that both
$body(r_1)$ and $body(r_2)$ are false w.r.t. $M$. Moreover, it holds that for each $M
\in SM(\bar{C} \cup Q)$, $\bar{I} \subseteq M$, as $Q=\{a\leftarrow \mid a \in
\bar{I}\}$. As a consequence, $body(r_2)$ is false w.r.t. $\bar{I}$.

Since the elements occurring in $body(r_2)$ are labeled literals, we have
proven that $cond(s) = [\gp_k]$ and that the condition $\forall a \in
content(s),$ $a$ is true for $\gp_k$ w.r.t. $\bar{I}$ does not hold. Such a
result, according to Definition~\ref{def:SC_sat} (page~\pageref{def:SC_sat}),
contradicts the hypothesis that $s$ is true for $\gp_j$ w.r.t. $\bar{I}$ and,
therefore, concludes the proof of the basis of the induction, in case (1).

In case (2), since $skel(s)=\emptyset$, according to
Definition~\ref{def:guess&check} (page~\pageref{def:guess&check}), there exists
a set of rules $S=\{r_1\} \cup \{s_i \mid 1 \leq i \neq j \leq n\} \subseteq
(\bar{C} \cup Q)$ such that:

\begin{center}
$ {\small
\begin{array}{lrl}

r_1: & \rho(s)_{\gp_j} \leftarrow & l \leq \mathtt{\#count}\{ K:
                                  (g(s))(K)_{\gp_j}, K \neq j\} \leq h \\
s_i: & (g(s))(i)_{\gp_j} \leftarrow & \bigwedge_{b \in \mathit{content}(s)} b_{\gp_i}
\qquad \qquad \qquad \qquad \qquad \qquad (1 \leq i \neq j \leq n)\mbox{.}
\end{array}
} $
\end{center}

Moreover, for each $r \in (\bar{C} \cup Q) \setminus S$ and for each $t \in S$,
$head(r) \neq head(t)$.

Now, since $\rho(s)_{\gp_j} \not\in M$, $body(r_1)$ is false w.r.t. $M$. Since
$body(r_1)= l \leq \mathtt{\#count}\{ K: (g(s))(K)_{\gp_j}, K \neq j\} \leq h$,
according to the definition of aggregate functions~\cite{DLPA03}, for each $D
\subseteq \{i \mid 1 \leq i \neq j \leq n \}$ ($l \leq |D| \leq h$), there exists $k$
s.t. $1 \leq k \neq j \leq n$ and $(g(s))(k)_{\gp_j}$ is false w.r.t. $M$. Thus,
there exists a rule $s_k \in \bar{C} \cup Q$ s.t. $s_k:  (g(s))(k)_{\gp_j} \leftarrow
\bigwedge_{b \in \mathit{content}(s)} b_{\gp_k}$ and $head(s_k)$ is false w.r.t. $M$.
Since $\forall r \in (\bar{C} \cup Q)\setminus \{s_k\}$, $head(r) \neq head((s_k)$,
$body(s_k)$ is false w.r.t. $M$, i.e. $\bigwedge_{b \in \mathit{content}(s)}
b_{\gp_k}$ is false w.r.t. $M$. Now, since $\bar{I} \subseteq M$, $\bigwedge_{b \in
\mathit{content}(s)} b_{\gp_k}$ is false w.r.t. $\bar{I}$.

Thus, we have proven that for each set $D \subseteq \{ i \mid 1 \leq i \neq j
\leq n\}$ s.t. $l \leq |D| \leq h,$ there exists some $k \in D$ and some $x \in
\mathit{content}(s)$ s.t. $x$ is false for $\gp_k$ w.r.t. $\bar{I}$. This
result, according to item (2) of Definition~\ref{def:SC_sat}
(page~\pageref{def:SC_sat}), contradicts the hypothesis that $s$ is true for
$\gp_k$ w.r.t. $\bar{I}$. Now we have concluded the proof of of the basis of
the induction.

({\em Induction}). Assume that the statement holds for $max_{s' \in
N(s)}\{depth(s')\}=d>0$ and consider the case $max_{s' \in
N(s)}\{depth(s')\}=d+1$. First, observe that  $s$ is not simple, because
$skel(s) \neq \emptyset$. Since $s$ is well-formed, $cond(s)=[l,h]$. Thus,
according to Definition~\ref{def:guess&check} (page~\pageref{def:guess&check}),
there exists a set of rules $S=\{r_1\} \cup \{s_i \mid 1 \leq i \neq j \leq n\}
\subseteq (\bar{C} \cup Q)$ s.t.:

\begin{center}
$ {\small
\begin{array}{lrl}

r_1: & \rho(s)_{\gp_j} \leftarrow & l \leq \mathtt{\#count}\{ K:
                                  (g(s))(K)_{\gp_j}, K \neq j\} \leq h \\
s_i: & (g(s))(i)_{\gp_j} \leftarrow & \bigwedge_{b \in \mathit{content}(s)} b_{\gp_i}
    \ \And \ \bigwedge_{s' \in skel(s)} \rho(s')_{\gp_j} \qquad \qquad \qquad (1 \leq i \neq j \leq n)\\
\end{array}
} $
\end{center}

Now, observe that:
\begin{description}
\item (1) For each $s' \in skel(s)$, $max_{\sigma \in N(s')}\{depth(\sigma)\}=d$, and

\item (2) Since $s$ is true for $P_k$ w.r.t. $\bar{I}$, according to
          Definition~\ref{def:SC_sat} (page~\pageref{def:SC_sat}), for each $s' \in skel(s)$, $s'$ is true
          for $\gp_j$ w.r.t. $\bar{I}$.
\end{description}

On the basis of the above observations and the induction hypothesis, it holds that
for each $s' \in skel(s)$, $\rho(s')_{\gp_j} \in M$.

Since $s$ is true for $\gp_j$ w.r.t. $\bar{I}$ and $ \mathit{cond}(s) = [l,h]$,
according to Definition~\ref{def:SC_sat} (page~\pageref{def:SC_sat}) there
exists $D \subseteq SP \setminus \{\gp_j\}$ s.t. $l \leq |D| \leq h$ and
$\forall a \in \mathit{content}(s),$ $\forall \gp \in D,$ $a \mbox{ is true for
$\gp$ w.r.t. } \bar{I}$.

Thus, it holds that, for each $1 \leq i \leq n$ such that both $i \neq j $ and $\gp_i
\in D$, $\bigwedge_{b \in \mathit{content}(s)} b_{\gp_i}
    \ \And \ \bigwedge_{s' \in skel(s)} \rho(s')_{\gp_j}$ is true w.r.t. $M$,
as $\bar{I} \subseteq M$. Since $M$ is a stable model of $\bar{C} \cup Q$, according
to the definition of the set $S$ (given at the beginning of the part ({\em
Induction})), for each $i$ s.t. $1 \leq i \neq j \leq n$ and s.t. $\gp_i \in D$,
$(g(s))(i)_{\gp_j}$ is true w.r.t. $M$.

Since $l \leq |D| \leq h$, there exists a set of literals $D'=\{ (g(s))(i)_{\gp_j}
\mid 1 \leq i \neq j \leq n\}$ s.t. $l \leq |D'| \leq h$ and s.t. for each element $d
\in D'$, $d$ is true w.r.t. $M$.

Now, according to the definition of aggregate functions~\cite{DLPA03}, $body(r_1)$ is
true w.r.t. $M$ and, since $M$ is a stable model of $\bar{C} \cup Q$, $head(r_1)$ is
true w.r.t. $M$, i.e. $\rho(s)_{\gp_j}$ is true w.r.t. $M$. Such a result concludes
the {\em only-if} part ($\Rightarrow$) of the proof.

($\Leftarrow$). We have to prove that if $\exists M \in SM(\bar{C} \cup Q) \mid
\rho(s)_{\gp_j} \in M$, then $s$ is true for $\gp_j$ w.r.t. $\bar{I}$. We proceed by
induction on the maximum nesting depth of the elements in $N(s)$, $d=max_{s' \in
N(s)}\{depth(s')\}$ $(d \geq 0)$.

({\em Basis}). In case $d=0$, then $N(s)=\{s\}$ and $depth(s)=0$. Since $|N(s)|=1$,
$s$ is a simple SC, i.e. $skel(s)=\emptyset$. Observe now that it may occur either
the following cases: (1) $cond(s)=[\gp_k]  (1 \leq k \leq n)$, or (2)
$cond(s)=[l,h]$.

In case (1), according to Definition~\ref{def:guess&check}
(page~\pageref{def:guess&check}), there exist two rules in $\bar{C} \cup Q$ of
the form:

\begin{center}
$ {\small
\begin{array}{lrl}

r_1: & \rho(s)_{\gp_j} \leftarrow & (g(s))(k)_{\gp_j}\\
r_2: & (g(s))(k)_{\gp_j} \leftarrow & \bigwedge_{b \in \mathit{content}(s)} b_{\gp_k}\\

\end{array}
} $
\end{center}

\noindent and such that $\forall r \in (\bar{C} \cup Q) \setminus \{r_1, r_2\}$,
$head(r) \neq head(r_1) \ \And \ head(r) \neq head(r_2)$.

As a consequence, since $\rho(s)_{\gp_j} \in M$, $body(r_1)$ is true w.r.t. $M$. Now,
since $M$ is a stable model of $\bar{C} \cup Q$, $body(r_2)$ is true w.r.t. $M$.

Now, observe that according to Definitions~\ref{def:rho&g}
and~\ref{def:guess&check} (see pages~\pageref{def:rho&g}
and~\pageref{def:guess&check}), for each $b \in content(s)$, $b_{\gp_k} \not\in
(M \setminus \bar{I})$, because $M \setminus \bar{I}$ includes only literals
that are auxiliary to the translation. As a consequence, $body(r_2)$ is true
w.r.t. $\bar{I}$, i.e. for each $b \in content(s)$, $b$ is true for $\gp_k$
w.r.t. $\bar{I}$.

Now, we have proven that $cond(s)=[\gp_k] (1 \leq k \leq n)$ and there exists
$\gp_k \in SP$ s.t. for each $a \in content(s)$, $a$ is true for $\gp_k$ w.r.t.
$\bar{I}$, thus, $s$ is true for $\gp_k$ w.r.t. $\bar{I}$. This concludes the
proof of the basis of the induction, in case (1).

In case (2), since $skel(s)=\emptyset$, according to
Definition~\ref{def:guess&check} (page~\pageref{def:guess&check}), there exists
a set of rules $S=\{r_1\} \cup \{s_i \mid 1 \leq i \neq j \leq n\} \subseteq
(\bar{C} \cup Q)$ such that:

\begin{center}
$ {\small
\begin{array}{lrl}

r_1: & \rho(s)_{\gp_j} \leftarrow & l \leq \mathtt{\#count}\{ K:
                                  (g(s))(K)_{\gp_j}, K \neq j\} \leq h \\
s_i: & (g(s))(i)_{\gp_j} \leftarrow & \bigwedge_{b \in \mathit{content}(s)} b_{\gp_i}
\qquad \qquad \qquad \qquad \qquad \qquad (1 \leq i \neq j \leq n)\mbox{.}
\end{array}
} $
\end{center}

Moreover, for each $r \in (\bar{C} \cup Q) \setminus S$ and for each $t \in S$,
$head(r) \neq head(t)$.

Since $M$ is a stable model of $\bar{C} \cup Q$ and $\rho(s)_{\gp_j} \in M$,
$body(r_1)$ is true w.r.t. $M$. According to the definition of aggregate
functions~\cite{DLPA03}, there exists a set of integers $D$ s.t. $l \leq |D|
\leq h$, for each $i$, $i \neq j$ and $\gp_i \in SP$ and, finally, for each $i
\in D$, $(g(s))(i)_{\gp_j}$ is true w.r.t. $M$.

As a consequence, there exists a set of rules $D'=\{ s_i \mid i \in D \ \And \
head(s_i) = (g(s))(i)_{\gp_j} \} \subseteq S \setminus \{r_1\}$ s.t. for each
rule $s_i \in D'$, $body(s_i)$ is true w.r.t. $M$. Since $\bar{I} \subseteq M$,
it holds that for each $s_i \in D'$, $body(s_i)$ is true w.r.t. $\bar{I}$. Now,
observe that $l \leq |D'|=|D| \leq h$ and that, according to
Definition~\ref{def:guess&check} (page~\pageref{def:guess&check}), for each
$s_i \in D$, $body(s_i) = \bigwedge_{b \in \mathit{content}(s)} b_{\gp_i}$. We
have obtained that there exists a set $D$ s.t. $l \leq |D| \leq h$, for each $i
\in D$, $i \neq j$ and $\gp_i \in SP$. Finally, it holds that for each $b \in
content(s)$ and for each $i \in D$, $b$ is true for $\gp_i$ w.r.t. $\bar{I}$.
Now, it results that there exists a set $\Delta=\{ \gp_i \mid i \in D \ \And \
\gp_i \in SP \}$ s.t. $\Delta \subseteq SP\setminus \{\gp_j\}$, $l \leq |Delta|
= |D| \leq h$ and for each $b \in content(s)$, and for each $\gp \in \Delta$,
$b$ is true for $\gp_i$ w.r.t. $\bar{I}$.

According to Definition~\ref{def:SC_sat} (page~\pageref{def:SC_sat}), we have
proven that $s$ is true for $\gp_j$ w.r.t. $\bar{I}$. Such a result concludes
the proof of the basis of the induction.

({\em Induction}). Assume that the statement holds for $max_{s' \in
N(s)}\{depth(s')\}=d >0$. and consider the case $max_{s' \in
N(s)}\{depth(s')\}=d+1$. First, observe that  $s$ is not simple. Since $s$ is
well-formed, $cond(s)=[l,h]$. Thus, according to
Definition~\ref{def:guess&check} (page~\pageref{def:guess&check}), there exists
a set of rules $S=\{r_1\} \cup \{s_i \mid 1 \leq i \neq j \leq n\} \subseteq
(\bar{C} \cup Q)$ s.t.:

\begin{center}
$ {\small
\begin{array}{lrl}

r_1: & \rho(s)_{\gp_j} \leftarrow & l \leq \mathtt{\#count}\{ K:
                                  (g(s))(K)_{\gp_j}, K \neq j\} \leq h \\
s_i: & (g(s))(i)_{\gp_j} \leftarrow & \bigwedge_{b \in \mathit{content}(s)} b_{\gp_i}
    \ \And \ \bigwedge_{s' \in skel(s)} \rho(s')_{\gp_j} \qquad \qquad \qquad (1 \leq i \neq j \leq n)\\
\end{array}
} $
\end{center}

Moreover, for each $r \in (\bar{C} \cup Q) \setminus S$ and for each $t \in S$,
$head(r) \neq head(t)$.

Now, observe that, since $\rho(s)_{\gp_j} \in M$ and $M$ is a stable model of
$\bar{C} \cup Q$, according to the definition of aggregate functions~\cite{DLPA03},
there exists a set $\Delta' \subseteq \{(g(s))(i)_{\gp_j} \mid 1 \leq i \neq j \leq
n\}$ s.t. $l \leq |\Delta'| \leq h$ and $\forall x \in \Delta'$, $x$ is true w.r.t.
$M$. Thus, there exists $\Delta'' \subseteq S \setminus \{r_1\}$ s.t. $l \leq
|\Delta''| = |\Delta'| = |\Delta| \leq h$ and s.t. for each $s_i \in \Delta''$,
$head(s_i)$ is true w.r.t. $M$. Since $M$ is a stable model of $\bar{C} \cup Q$, for
each $s_i \in \Delta''$, $body(s_i)$ is true w.r.t. $M$. Now, note that for each $i$
s.t. $s_i \in \Delta''$, $body(s_i)=\bigwedge_{b \in \mathit{content}(s)} b_{\gp_i}
    \ \And \ \bigwedge_{s' \in skel(s)} \rho(s')_{\gp_j}$.

Thus, for each $i$ s.t. $s_i \in \Delta''$, $\bigwedge_{b \in \mathit{content}(s)}
b_{\gp_i}$ is true w.r.t. $M$ and $\bigwedge_{s' \in skel(s)} \rho(s')_{\gp_j}$ is
true w.r.t. $M$.

Now, since $\bar{I} \subseteq M$ and according to
Definition~\ref{def:guess&check} (page~\pageref{def:guess&check}), for each $i$
s.t. $s_i \in \Delta''$, $\bigwedge_{b \in \mathit{content}(s)} b_{\gp_i}$ is
true w.r.t. $\bar{I}$. Since for each $i$ s.t. $s_i \in \Delta''$, $b_{\gp_i}$
is a literal labelled w.r.t. $\gp_i$, it is easy to see that for each $i$ s.t.
$s_i \in \Delta''$ and for each $b \in \mathit{content}(s)$, $b$ is true for
$\gp_i$ w.r.t. $\bar{I}$. Moreover, by induction hypothesis, for each $s' \in
skel(s)$, $s'$ is true for $\gp_j$ w.r.t. $\bar{I}$.

Thus, we have obtained that $cond(s)=[l,h]$ and there exists a set $D \subseteq
\{i \mid 1 \leq i \neq j \leq n \}$ s.t. $l \leq |D| \leq h$ and for each $i
\in D$ s.t. $\gp_i \in SP$ and for each $a \in content(s)$, it holds that $a$
is true w.r.t. $\gp$ and for each $s' \in skel(s)$, $s'$ is true for $\gp_j$
w.r.t. $\bar{I}$. According to Definition~\ref{def:SC_sat}
(page~\pageref{def:SC_sat}), we have proven that $s$ is true for $\gp_j$ w.r.t.
$\bar{I}$. This concludes the proof of the lemma.
\end{proof}

\begin{thm*}{Theorem}{the:1to1correspndence}
\maintheorem
\end{thm*}

\begin{proof}
Before starting with the proof, let us denote $C(\gp_1, \cdots, \gp_n)$ by $\bar{C}$.

($\subseteq$). By contradiction, assume that $\forall X \in A, X \not\in B $. Observe
that, according to Definitions~\ref{def:SCtrans},~\ref{def:P'_u} and~\ref{def:[]}
(see pages~\pageref{def:SCtrans},~\pageref{def:P'_u} and~\pageref{def:[]},
respectively), it holds that $X=\bar{F} \cup \bar{G} \cup \bar{H}$ such that
$\bar{F}\cap\bar{G}=\emptyset$, $\bar{G}\cap\bar{H}=\emptyset$,
$\bar{F}\cap\bar{H}=\emptyset$. Thus, we prove that either condition (1), (2), or (3)
is false. We consider each condition separately.

\paragraph{Condition (1).}

In case condition (1) is false, it holds that $\bar{F} \not\in
\mathcal{SOS}(\gp_1, \cdots, \gp_n)$. It follows that either:
\begin{description}
\item[] ($a$) $\exists i \mid (F^i)_{\gp_i} \subseteq \bar{F} \ \And \ F^i \not\in
AFP(\gp_i)$, or

\item[] ($b$) $\exists i,r \mid (F^i)_{\gp_i} \subseteq \bar{F} \ \And \ F^i \in
AFP(\gp_i) \ \And \ r \in \gp_i \ \And \ ST_C(\bar{F}) \neq \bar{F}$.

\end{description}

In case ($a$), by virtue of Lemma~\ref{pro:SM(Pu)_eq_union}
(page~\pageref{pro:SM(Pu)_eq_union}), it holds that $(F^i)_{\gp_i} \not\subseteq
SM(P_u)$. As a consequence, it is easy to see that $(F^i)_{\gp_i} \not\subseteq
SM(P'_u \cup \bar{C})$. Since $(F^i)_{\gp_i} \in X$, then we have thus reached a
contradiction.

If item ($b$) occurs, it holds that either:
\begin{description}
\item[] ($b_1$) $\exists a, \gp_i \mid \gp_i \in C \ \And \ a_{\gp_i} \in \bar{F} \
\And \
         a_{\gp_i} \not\in ST_C(\bar{F})$, or

\item[] ($b_2$) $\exists a, \gp_i \mid \gp_i \in C \ \And \ a_{\gp_i} \not\in \bar{F}
\ \And \
         a_{\gp_i} \in ST_C(\bar{F})$.
\end{description}

In case ($b_1$), both the following conditions are true:
\begin{description}
\item[] (i) $\forall r \in \gp_i, \mathit{body}(r)$ is true w.r.t. $\bar{F}$
            $\Rightarrow \mathit{head}(r) \neq a$, and

\item[] (ii) $\forall r \in \gp_i, \mathit{body}(r)$ is true w.r.t. $\bar{F} \ \And \
        a$ is true for $\gp_j$ w.r.t. $\bar{F}$
        $\Rightarrow \mathit{head}(r) \neq \mathit{okay}(a)$.
\end{description}

In case (i) of item ($b_1$), it holds that, according to
Definitions~\ref{def:SOLP_hat} and~\ref{def:gamma'P} (pages
~\pageref{def:SOLP_hat} and~\pageref{def:gamma'P}), for each $r' \in
S'_2(\hat{\gp_i})$, s.t. $\mathit{body}(r')$ is true w.r.t. $X$,
 it results that $\mathit{head}(r') \neq sa_{\gp_i}$. Thus, $sa_{\gp_i} \not\in X$.
As a consequence, according to the definition of $S'_3(\hat{\gp_i})$ (see
Definition~\ref{def:gamma'P}, page~\pageref{def:gamma'P}), it holds that
$a_{\gp_i} \not\in X$. Since the hypothesis requires that $a_{\gp_i} \in
\bar{F}$ and $\bar{F} \subseteq X$, we have reached a contradiction.

Consider now case (ii) of item ($b_1$). It holds that, according to
Definitions~\ref{def:SOLP_hat} and~\ref{def:gamma'P} (pages
~\pageref{def:SOLP_hat} and~\pageref{def:gamma'P}), for each $r' \in
S'_2(\hat{\gp_i})$ s.t. $a_{\gp_i} \ \And \ \mathit{body}(r')$ is true w.r.t.
$X$, it results that $\mathit{head}(r') \neq sa_{\gp_i}$. Thus, $sa_{\gp_i}
\not\in X$. According to the definition of $S'_3(\hat{\gp_i})$, it holds that
$a_{\gp_i} \not\in X$. Since the hypothesis requires that $a_{\gp_i}  \in
\bar{F}$ and $\bar{F} \subseteq X$, we have reached a contradiction. This
concludes the part of the proof concerning item ($b1$) above.

Consider now item ($b_2$). In this case at least one of the following conditions
holds:
\begin{description}
\item[] (i) $\exists r \in \gp_i \mid \mathit{body}(r)$ is true w.r.t. $\bar{F} \
\And \
        a$ is true for $\gp_j$ w.r.t. $\bar{F}
        \ \And \ \mathit{head}(r)=\mathit{okay}(a)$,

\item[] (ii) $\exists r \in \gp_i \mid \mathit{body}(r)$ is true w.r.t. $\bar{F} \
\And \
             \mathit{head}(r)=a$.
\end{description}

If case (i) of item ($b_2$) occurs, it holds that $a_{\gp_i} \in \bar{F}$. Now, the
contradiction is thus reached, since according to the hypothesis, $a_{\gp_i} \not\in
\bar{F}$.

Consider now case (ii) of item ($b_2$). Let $r$ be of the form $a \leftarrow
b_1, \cdots b_\nu, s_1, \cdots, s_m$ (we do not lose in generality because $r$
is any social rule). According to Definitions~\ref{def:SOLP_hat}
and~\ref{def:gamma'P} (pages ~\pageref{def:SOLP_hat}
and~\pageref{def:gamma'P}), there exists a rule $r' \in S'_2(\hat{\gp_i})$ such
that $r'$ has the form $sa_{\gp_i} \leftarrow b_{\gp_i}^1 , \cdots ,
b_{\gp_i}^\nu, \rho(s_1)_{\gp_i}, \cdots \rho(s_m)_{\gp_i}$.

Now, since $\mathit{body}(r)$ is true w.r.t. $\bar{F}$ and on the basis of
results of Lemma~\ref{lem:SC true iff rho in M} (page~\pageref{lem:SC true iff
rho in M}),
 it holds that $\mathit{body}(r')$ is true w.r.t. $X$
and $\mathit{head}(r')=sa_{\gp_i}$. According to the definition of
$S'_3(\hat{\gp_i})$ (see Definition~\ref{def:gamma'P},
page~\pageref{def:gamma'P}), since  $a_{\gp_i} \not\in X$, then it holds that
$sa_{\gp_i} \not\in X$. Thus, $\mathit{body}(r')$ is true w.r.t. $X$ and
$\mathit{head}(r')$ is false w.r.t. $X$. As a consequence, there exists a rule
$r'$ in $P'_u \cup \bar{C}$ such that $r'$ is false w.r.t. the model $X$ ($X
\in SM(P'_u \cup \bar{C})$) which is a contradiction. The proof of condition
(1) is thus concluded. Let us prove now condition (2).

\paragraph{Condition (2).}
In case condition (2) is false, there exists $i$ s.t. $(G^i)_{\gp_i} \neq
[F^i]_{\gp_i} \setminus (F^i)_{\gp_i}$. Thus, $(F^i)_{\gp_i} \cup (G^i)_{\gp_i} \neq
[F^i]_{\gp_i}$ and then, according to Definition~\ref{def:[]}
(page~\pageref{def:[]}), $(F^i)_{\gp_i} \cup (G^i)_{\gp_i} \not\subseteq \bigcup_{F
\in AFP(\gp_i)}\{[F]_{\gp_i}\}$. Now, by virtue of Lemmas~\ref{pro:SM_eq_union}
and~\ref{pro:SM(Pu)_eq_union} (pages~\pageref{pro:SM_eq_union}
and~\pageref{pro:SM(Pu)_eq_union}), $(F^i)_{\gp_i} \cup (G^i)_{\gp_i} \not\subseteq
SM(P_u)$ (recall that $P_u = \bigcup_{1 \leq i \leq n}\Gamma'(A(\hat{\gp_i}))$).
According to Definitions~\ref{def:SCtrans},~\ref{def:gamma'P} and~\ref{def:P'_u} (see
pages~\pageref{def:SCtrans},~\pageref{def:gamma'P} and~\pageref{def:P'_u},
respectively), it is easy to see that $(F^i)_{\gp_i} \cup (G^i)_{\gp_i} \not\subseteq
SM(P'_u \cup \bar{C})$. Since $(F^i)_{\gp_i} \cup (G^i)_{\gp_i} \in X$, we have
reached a contradiction. Consider now the last case.

\paragraph{Condition (3).}
If condition (3) is false, then there exists $i$ such that $H_{\gp_i}^i$ is not
equal to $\bigcup_{s \in MSC^{\gp_i}} SAT^{\gp_i}_{\bar{F}}(s)$. Thus, it holds
that either:

\begin{description}

\item[] ($a$) $\exists h \in (H^i)_{\gp_i} \mid h \not\in \bigcup_{s \in
        MSC^{\gp_i}} SAT^{\gp_i}_{\bar{F}}(s)$, or

\item[] ($b$) $\exists h \in \bigcup_{s \in MSC^{\gp_i}} SAT^{\gp_i}_{\bar{F}}(s)
        \mid h \not\in (H^i)_{\gp_i}$.
\end{description}

In case ($a$), according to Definition~\ref{def:SAT} (page~\pageref{def:SAT}),
since $h \not\in \bigcup_{s \in MSC^{\gp_i}} SAT^{\gp_i}_{\bar{F}}(s)$, it
holds that $\forall M \in SM(\bar{C} \cup Q)$ s.t. $Q=\{a\leftarrow \mid a \in
\bar{F}\}$, it results that $h \not\in M$. As a consequence, according to
Definition~\ref{def:P'_u} (page~\pageref{def:P'_u}), we have that $\forall M
\in SM(P'_u \cup \bar{C})$, $h \not\in M$. Now, we
have reached a contradiction.\\

In case ($b$), according to Definition~\ref{def:SAT} (page~\pageref{def:SAT}), $h$ is
either an auxiliary $\rho$-atom or a $g$-predicate and it holds that $h \not\in
(H^i)_{\gp_i}$. Since $\bar{H} = \bigcup_{1 \leq k \leq n} (H^k)_{\gp_k}$, and $h$ is
labeled w.r.t. $\gp_i$, $h \not\in \bar{H}$.

Now we have reached a contradiction, because $h \in X \setminus \bar{H} = \bar{F}
\cup \bar{G}$ and $\bar{F} \cup \bar{G}$ does not include, according to conditions
(1) and (2) of the theorem statement, either $\rho$-atoms or $g$-predicates.

($\supseteq$). By contradiction, assume that $\forall X \in B, X= \bar{F} \cup
\bar{G} \cup \bar{H}$ and both conditions (1), (2) and (3) of the theorem statement
hold and further $X \not\in A$ (recall that $A=SM(P'_u \cup \bar{C}))$.

As a consequence, either:

\begin{description}
\item[] ($a$) $\exists r \in P'_u \cup \bar{C} \mid r$ is false w.r.t. $X$, i.e.
        $\mathit{head}(r)$ is false w.r.t. $X$ and $\mathit{body}(r)$ is true w.r.t. $X$, or

\item[] ($b$) $\exists X' \subset X \mid \forall r \in P'_u \cup \bar{C}, r$ is true
        w.r.t. $X'$.
\end{description}

In case ($a$), according to Definition~\ref{def:gamma'P}
(page~\pageref{def:gamma'P}), there exist $\gp_i,r'$ s.t. $\gp_i \in SP$, $r'
\in \gp_i$, $r'$ is a social rule $a \leftarrow b_1, \cdots b_\nu,$ $s_1,
\cdots, s_m$ and $r$, according to the definition of $S_2'(\hat{\gp_i})$, has
the form $sa_{\gp_i} \leftarrow b_{\gp_i}^1 , \cdots , b_{\gp_i}^\nu,$ $
\rho(s_1)_{\gp_i}, \cdots \rho(s_m)_{\gp_i}$.

Now, since $\mathit{body}(r)$ is true w.r.t. $X$ and $X= \bar{F} \cup \bar{G} \cup
\bar{H}$, for each $k$ ($1 \leq k \leq \nu$), $b_{\gp_i}^k$ is true w.r.t. $\bar{F}$.
Therefore, for each $k$ ($1 \leq k \leq \nu$), $b$ is true for $\gp_i$ w.r.t.
$\bar{F}$.

By virtue of Lemma~\ref{lem:SC true iff rho in M} (page~\pageref{lem:SC true
iff rho in M}), for each $l$ ($1 \leq l \leq m$), $s_l$ is true for $\gp_i$
w.r.t. $\bar{F}$. Thus, it holds that $\mathit{body}(r')$ is true w.r.t.
$\bar{F}$. Moreover, according to the definition of $S'_3(\gp_i)$ (see
Definition~\ref{def:gamma'P}, page~\pageref{def:gamma'P}), $\mathit{head}(r')$
is false w.r.t. $\bar{F}$, since $\mathit{head}(r)$ is false w.r.t. $\bar{F}$.

 As a consequence, the social rule $r' \in \gp_i$ is false w.r.t. $\bar{F}$.
Therefore, $\bar{F}$ is not a social model of $\gp_1, \cdots, \gp_n$. Now, we have
reached a contradiction.

In case ($b$), at least one of the following conditions holds, either:

\begin{description}
\item[] ($\alpha$) $(X \setminus X') \cap \bar{F} \neq \emptyset$, or

\item[] ($\beta$) $(X \setminus X') \cap \bar{G} \neq \emptyset$, or

\item[] ($\gamma$) $(X \setminus X') \cap \bar{H} \neq \emptyset$.
\end{description}

If condition ($\alpha$) occurs, according to both the hypothesis and
Definition~\ref{def:gamma'P} (page~\pageref{def:gamma'P}), then there exist
$\gp_i,r,a$ s.t $\gp_i \in SP, r \in S'_1(\gp_i), a_{\gp_i} \in (X \setminus
X') \cap \bar{F},$ and $r$ has the form $a'_{\gp_i} \leftarrow \naf a_{\gp_i}$.
Now, since $a_{\gp_i} \in (X \setminus X') \cap \bar{F}$, it holds that
$a_{\gp_i} \in X$ and $a'_{\gp_i} \not\in X$ (otherwise, according to $r$, $X$
would not be a model of $P'_u \cup \bar{C}$). Now, since $X' \subseteq X$, it
holds that $a'_{\gp_i} \not\in X'$. Therefore, there exists a social rule $r
\in P'_u \cup \bar{C}$
 s.t. $r$ is false w.r.t. $X'$. As a consequence, $X'$ is not a model of $P'_u \cup \bar{C}$
 and we have reached a contradiction.
This concludes the proof of case ($b$), condition ($\alpha$).

In case ($b$), if condition ($\beta$) holds, then according to both the
hypothesis and Definition~\ref{def:gamma'P} (page~\pageref{def:gamma'P}),
either:

\begin{description}

\item[] (i) There exist $\gp_i,r,a'$ s.t. $\gp_i \in SP$, $r \in S'_1(\gp_i)$,
$a'_{\gp_i} \in
            (X \setminus X') \cap \bar{G}$
            and $r$ has the form $a_{\gp_i} \leftarrow \naf a'_{\gp_i}$, or

\item[] (ii) There exist $\gp_i,r,a$ s.t. $\gp_i \in SP$, $r \in S'_3(\gp_i)$,
$a_{\gp_i} \in \bar{X}$,
        $sa_{\gp_i} \in (X \setminus X') \cap \bar{G}$ and $r$ has the form
        $fail_{\gp_i} \leftarrow \naf fail_{\gp_i}, a_{\gp_i}, \naf sa_{\gp_i}$.
\end{description}

If item (i) is true, then $r$ is false w.r.t. $X'$, since $a'_{\gp_i} \not\in X'$ and
$a_{\gp_i} \not\in X'$.
 Thus, $X'$ is not a model of $P'_u \cup \bar{C}$ and we have reached a contradiction.

If item (ii) holds, then $r$ is false w.r.t. $X'$, since, according to the
hypothesis, $sa_{\gp_i} \in X$. Therefore, $a_{\gp_i} \in X$. As a result,
$sa_{\gp_i} \not\in X'$ and $a_{\gp_i} \in X'$. Thus, $X'$ is not a model of of $P'_u
\cup \bar{C}$ and we have reached a contradiction. This concludes the proof of case
($b$), condition ($\beta$). Now we give the proof when condition ($\gamma$) holds.

In case ($b$), if condition ($\gamma$) holds, then there exist $\gp_i, h$ s.t.
$\gp_i \in SP$ and $h_{\gp_i} \in (X \setminus X') \cap \bar{H}$. Since $X'$ is
a model of $P'_u \cup \bar{C}$, according to Definitions~\ref{def:rho&g}
and~\ref{def:SAT} (pages~\pageref{def:rho&g} and~\pageref{def:SAT}), there
exists some $a_{\gp_i}$ in $\bar{F} \cup \bar{G}$ s.t. $a_{\gp_i} \in (X
\setminus X')$. Therefore, either condition ($\alpha$) or condition ($\beta$)
of case ($b$) occurs and it is easy to see that we have reached a
contradiction. Now we have concluded the proof of the theorem.
\end{proof}

\Listofabbrev
\end{document}